\documentclass[12pt]{article}
\usepackage{url}
\usepackage{graphicx}
\usepackage{amsmath}
\usepackage{amssymb}
\usepackage{booktabs}
\usepackage{subcaption}
\usepackage[capitalise,noabbrev]{cleveref}
\usepackage{multicol}
\usepackage{siunitx}
\usepackage[table]{xcolor}
\usepackage{textcomp}
\usepackage{setspace}
\usepackage{geometry}
\usepackage{authblk}
\usepackage{xurl} % better URL line breaking
\newcommand{\WIDIM}{{\tt WIDIM}}

\newcommand{\LIMA}{{\tt LIMA}}
\newcommand{\BLIMA}{{\tt BLIMA}}

\newcommand{\RAFT}{{\tt RAFT}}

\newcommand{\rev}[1]{\textcolor{black}{#1}}   % revision highlight -- set body to {#1} to clear

\begin{document}

% TODO[claudio]: NatComm word limit on title is 15 words. Current draft = 13 words.
\title{Shallow may be better: Physics-Constrained Lightweight Networks for High-Throughput Particle Image Velocimetry}

\author{Claudio Mucignat$^{1,*}$, Kamila Zdybał$^{1}$, Ivan Lunati$^{1}$}
\date{}
\maketitle

\noindent$^1$Laboratory for Computational Engineering,
Swiss Federal Laboratories for Materials Science and Technology (Empa),
\"{U}berlandstrasse 129, D\"{u}bendorf 8600, Switzerland.\\
$^*$Correspondence: \texttt{claudio.mucignat@empa.ch}

\bigskip

%%%%%%%%%%%%%%%%%%%%%%%%%%%%%%%%%%%%%%%%%%%%%%%%%%%%%%%%%%%%%%%%%%%%%
%% ABSTRACT                                                         %
%% NatComm: <= 200 words, no references.                            %
%% Current draft: ~190 words.                                       %
%%%%%%%%%%%%%%%%%%%%%%%%%%%%%%%%%%%%%%%%%%%%%%%%%%%%%%%%%%%%%%%%%%%%%
\section*{Abstract}

%Convolutional networks for optical flow typically adopt progressively deeper pyramidal encoders under the assumption that larger receptive fields are universally beneficial. For particle image velocimetry (PIV), however, tracer-particle displacement between frames is bounded by the acquisition geometry to a few tens of pixels, allowing the useful network depth to be derived from the physically admissible displacement rather than inherited from general-purpose architectures. Using a layer-by-layer analysis of the spatial support growth, we derive an explicit upper bound and show that a deep architecture exceeds it by two orders of magnitude, destabilizing training and degrading spatial resolution. The resulting shallower architecture, \LIMA{$^{3}$}, is compact by design, with 589,000 trainable parameters. Across synthetic and experimental benchmarks, \LIMA{$^{3}$} matches established learned and classical baselines while scaling linearly in memory and sustaining 900 image pairs per second at $640\times480$ resolution on a consumer GPU or 100 on an edge computing device. More generally, deriving the network spatial support from bounded displacement provides a transferable design principle for compact optical-flow architectures, offering an alternative to compressing oversized general-purpose networks and enabling efficient deployment alongside experiments.

Convolutional networks for optical flow typically adopt progressively deeper pyramidal encoders under the assumption that larger receptive fields improve motion estimation. For particle image velocimetry (PIV), however, tracer-particle displacement between frames is bounded by the acquisition geometry, allowing network depth to be derived from the physically admissible displacement rather than inherited from general-purpose architectures. Using a layer-by-layer analysis of spatial-support growth, we derive its explicit upper bound and show that deeper architectures substantially exceed the support required for PIV, compromising training stability, spatial resolution under particle loss, and inference efficiency. Based on this analysis, we develop $\LIMA^{3}$, a lightweight pyramidal displacement estimator with 589,000 trainable parameters. Across synthetic and experimental benchmarks, $\LIMA^{3}$ matches the accuracy of established learned and classical methods with at least an order of magnitude less memory and substantially higher throughput, while retaining the physically expected coarsening of spatial resolution as tracer information is lost. Its linear memory scaling enables megapixel processing within embedded memory budgets, such that $\LIMA^{3}$ can reach over 900 image pairs per second at $640\times480$~pixel on an RTX 5090 and 99 image pairs per second on an NVIDIA Jetson AGX Orin. The spatial-support analysis provides a transferable design principle for compact optical-flow architectures and bounded-window attention, offering an alternative to compressing oversized general-purpose networks and enabling efficient operation alongside image acquisition.
%%%%%%%%%%%%%%%%%%%%%%%%%%%%%%%%%%%%%%%%%%%%%%%%%%%%%%%%%%%%%%%%%%%%%
%% INTRODUCTION                                                     %
%% NatComm: place the central claim early; broad readership.        %
%% Opens with the counterintuitive finding (moved from Conclusions  %
%% paragraph beginning "It is important to emphasize...") and       %
%% motivates it.                                                    %
%%%%%%%%%%%%%%%%%%%%%%%%%%%%%%%%%%%%%%%%%%%%%%%%%%%%%%%%%%%%%%%%%%%%%
\section*{Introduction}
\label{sec:intro}

Optical-flow architectures, inherited from natural-image computer vision, rest on the assumption that deeper pyramidal encoders and larger receptive fields are universally beneficial: deeper hierarchies capture richer correspondences, while longer-range context helps resolve matching ambiguities. We argue that this assumption is not generally justified when the useful network depth is constrained by the physical displacement scale of the measurement.

\begin{figure}[htbp!]
    \centering
    \includegraphics[width=\textwidth]{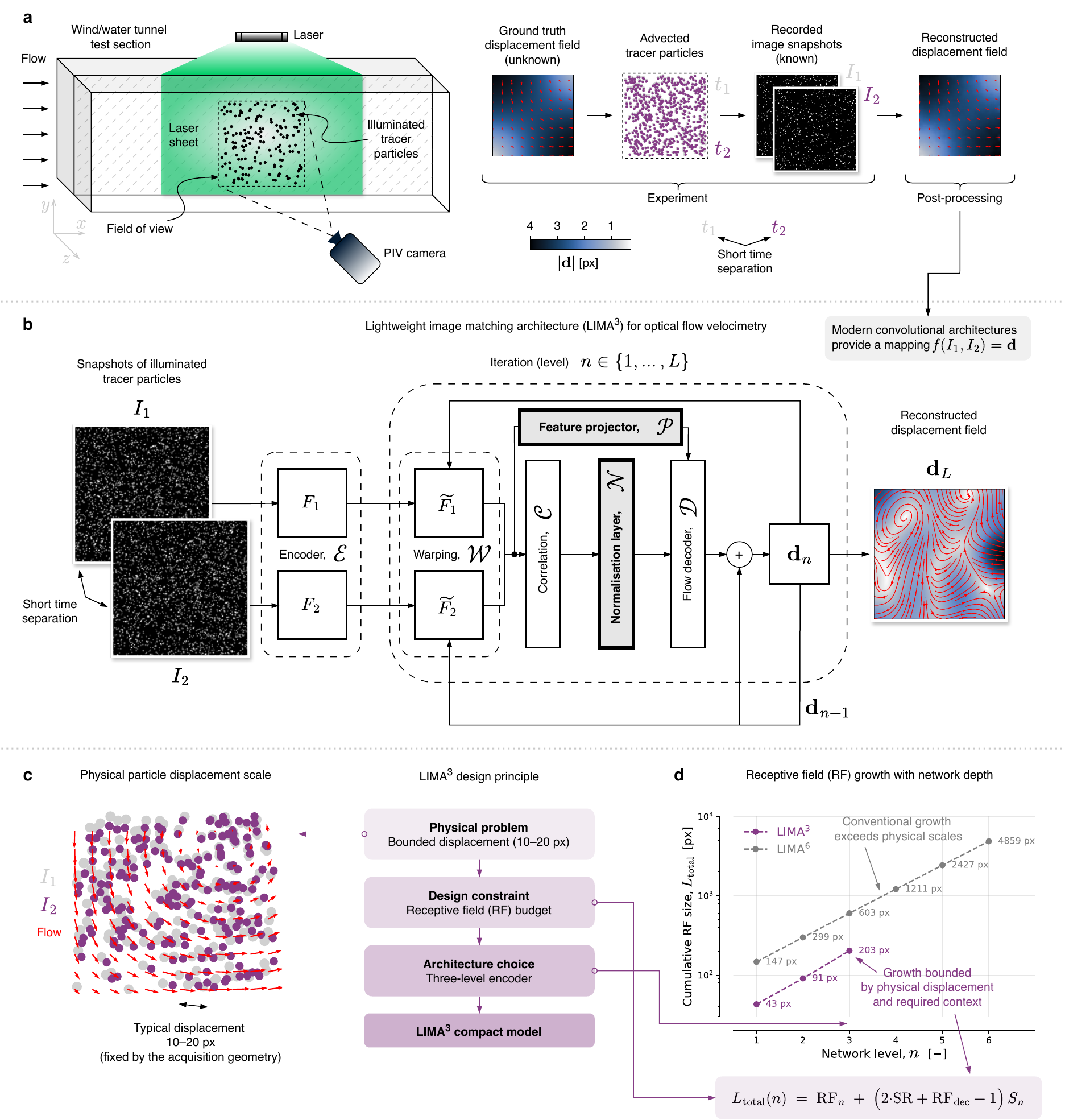}
    \caption{\textbf{Physics-constrained design of the lightweight image matching architecture ($\LIMA{^{3}}$) for particle image velocimetry (PIV).} \textbf{a} PIV experimental setup and post-processing. The unknown displacement field, $\mathbf{d}$, is estimated from recorded snapshots, $I_1$ and $I_2$, of illuminated tracer particles. \textbf{b} The proposed architecture comprises a Siamese feature encoder, coarse-to-fine warping and correlation, a flow decoder, and two lightweight stabilisation modules: correlation-volume normalisation, $\mathcal{N}$, and the two-channel feature projector, $\mathcal{P}$, highlighted in grey. \textbf{c} Inter-frame particle displacement is bounded by the acquisition geometry and typically lies within approximately 10--20 pixels; \textbf{d} The displacement bound constrains the required matching range and, consequently, the useful spatial-support budget. A three-level encoder provides sufficient spatial support for the targeted displacement regime while avoiding the unnecessarily large end-to-end receptive field, cumulative stride, and padding of deeper pyramids.}
    \label{fig:LIMA}
\end{figure}

An important example is dense displacement estimation for particle image velocimetry (PIV~\cite{Raffael2018}). PIV measures the motion of neutrally buoyant tracer particles between image pairs, as conceptually visualized in Figure~\ref{fig:LIMA}a, whose inter-frame displacement is limited by the acquisition geometry, typically to 10–20 pixels.
In this setting, a network whose deepest layer aggregates information across thousands of input pixels provides no useful long-range context. Instead, it integrates information from image regions that bear no physical relation to the displacement at the point of interest, while incurring unnecessary costs in memory and inference time.

We formalise this argument by analysing, layer by layer, how the receptive field grows with network depth and with the design of the encoder, correlation module, and decoder.  We compare the resulting spatial support with the displacement scales encountered in typical PIV experiments. The analysis shows that a six-level architecture, as used in many optical-flow architectures~\cite{sun2018pwc,hur2019iterative,teed2020raft}, has an end-to-end receptive field exceeding four thousand pixels in the input image, whereas the relevant particle displacements are typically only a few tens of pixels and must be measured with sub-pixels precision~\cite{Raffael2018}. 

Guided by this analysis, we design $\LIMA{^{3}}$, a compact pyramidal optical-flow architecture that builds on the lightweight image matching architecture ($\LIMA{}$, \cite{manickathan2023lightweight,mucignat2023lightweight,Mucignat2025}) but derives its encoder depth and decoder support from the spatial scales of the measurement, rather than assuming that progressively deeper pyramids and larger receptive fields necessarily improve correspondence estimation. $\LIMA{^{3}}$ retains three encoder levels, matches the end-to-end receptive field to the physically relevant correlation range, and incorporates two lightweight modules for stabilising the reconstruction: correlation-volume normalisation and a two-channel feature projector. 
Rather than compressing an oversized architecture or selecting its depth empirically by pruning, the design follows directly from the bounded displacement imposed by the measurement.

The resulting convolutional neural network (CNN) comprises nearly three times fewer parameters than the original $\LIMA{}$ architecture, approximately one order of magnitude fewer than the 
\RAFT{} family \cite{teed2020raft,lagemann2021deep,lagemann2022generalization}, and two orders of magnitude fewer than FlowNet2~\cite{ilg2017flownet}.  $\LIMA{^{3}}$ provides three output variants with different speed--accuracy--resolution trade-offs, denoted $\LIMA{^{3,0}}$, $\LIMA{^{3,1}}$, and $\LIMA{^{3,2}}$, which produce dense flow at full, half, and quarter resolution, respectively. This compact-by-design approach contrasts with the increasing complexity of optical-flow architecture. For instance, vision-transformer-based PIV networks, which adopt attention mechanisms developed for natural-image optical flow (including Twins-PIVNet~\cite{Reddy2025}, FlowFormer-PIV~\cite{Yu2026}, and the multi-cost-volume MCFormer~\cite{Lin2025}), contain one to two orders of magnitude more parameters than $\LIMA{^{3}}$. This raises the question of whether matching architectural complexity to the measurement scale can simultaneously reduce model complexity and improve reconstruction accuracy.

A practical challenge in comparing these approaches is that published networks are trained on different proprietary synthetic datasets, often generated with proprietary pipelines, with no common training distribution or image generator across the field. Architecture and training distribution are therefore confounded, and observed performance reflects both. Throughout the paper, our primary model is the deployed $\LIMA{^{3}}$, trained on our synthetic image generator, described in the Methods and made publicly available~\cite{Zdyba2025}. To disentangle architectural effects from training-distribution effects, we also include two variants of $\LIMA{^{3,0}}$ retrained on the two public \RAFT{} training datasets of~\cite{lagemann2021deep}: Problem~Class~1 and Problem~Class~2, the latter introduced in~\cite{lagemann2021deep} as a noisier variant of~\cite{Cai2019}. These variants serve as controls in the public-benchmark and resolution--noise analyses rather than as alternative deployment configurations.

We evaluate the trained architecture across three benchmark classes. First, a factorial ablation study isolates the effects of encoder depth, decoder receptive field, feature projection, and correlation normalisation. Comparing all combinations of the four design choices identifies the configuration adopted for $\LIMA{^3}$ and shows that six-level variants are not only larger but also \rev{require additional stabilisation of the coarsest levels to train}. Second, a sinusoidal spatial-resolution sweep under controlled out-of-plane particle loss maps all methods onto a resolution–noise plane, showing that the optimised three-level $\LIMA{}$ variants achieve the most favourable resolution–noise trade-off among the evaluated methods. The analysis further reveals that the increase in $\RAFT{}$ sharpness under particle loss arises from frequency locking rather than recovery of finer-scale structures. Third, we evaluate  $\LIMA{^{3}}$ on sixteen experimental configurations spanning diverse flow conditions and seven PIV Challenge cases~\cite{PIVChallenge}. Across these test cases,  $\LIMA{^{3,2}}$ processes 640×480-pixel frames at up to 900 image pairs per second on a single GPU or 100 image pairs per second on an edge computing device. Its low memory footprint and computational cost enable inference during experiments, concurrently with data acquisition rather than only during offline post-processing.

Together, these results support a general design principle that extends beyond PIV: for optical-flow problems with bounded displacement, network depth and spatial support should be matched to the characteristic scales of the measurement rather than maximised by default, improving reconstruction quality while reducing computational cost.

\section*{Results}
\label{sec:results}

We first establish the receptive-field bound used to set the architecture depth and show that three pyramid levels are sufficient for the displacement scales encountered in PIV. We then evaluate the resulting architecture through ablation experiments, benchmark flows, controlled particle-loss tests, experimental datasets, and computational measurements.
The derivation of the reported quantities is given in the Methods. Throughout, we compare $\LIMA{^3}$ variants with the window-deformation iterative multi-grid algorithm, \WIDIM{}~\cite{scarano2001iterative}, and the two $\RAFT$  variants available at~\cite{codeocean_RAFT}: \RAFT 
{$_{32}$}, which uses a denser feature representation to maximise spatial fidelity, and \RAFT 
{$_{256}$}, which uses a coarser representation to reduce memory and computational cost at the expense of spatial detail.  Details of the processing pipeline for \RAFT{} and \WIDIM{} are given in the Methods.

\subsection*{Receptive-field analysis and the upper bound on model spatial support}
\label{sec:rf_summary}

The full architecture of \LIMA{}, including the new lightweight modules for stabilization (correlation-volume normalisation and a two-channel feature projector), is illustrated in Figure~\ref{fig:LIMA}b. First, a Siamese encoder downsamples both frames into an $L$-level feature pyramid. At each level, a correlation operator, $\mathcal{C}$, compares the two feature maps within a fixed search range, $\mathrm{SR}$, and the result is fed into the normalisation layer, $\mathcal{N}$; a decoder, $\mathcal{D}$, converts the correlation responses into a displacement update with contribution from the two-channel feature projector, $\mathcal{P}$, and refines the estimate from coarse to fine. In a pyramidal correlation network, each output displacement is informed by a finite patch of the input image, which defines its end-to-end spatial support.

To determine the total spatial support, \textit{i.e.}, the cumulative receptive field $L_{\mathrm{total}}$, we combine the encoder's receptive field at level $n$, $\mathrm{RF}_n$, the correlation search range, $\mathrm{SR}$, and the decoder's receptive field, $\mathrm{RF}_{\mathrm{dec}}$, to obtain
\begin{equation}
L_{\mathrm{total}}(n) \;=\; \mathrm{RF}_n \;+\; \bigl(2 \cdot \mathrm{SR} + \mathrm{RF}_{\mathrm{dec}} - 1\bigr)\,S_n,
\label{eq:Ltot}
\end{equation}
where $S_n$ is the cumulative encoder stride at level $n$ (the full, layer-by-layer derivation is given in Methods). Because $S_n$ doubles at each level, the contribution of the correlation and decoder terms to $L_{\mathrm{total}}$ grows exponentially with depth. However, the physical displacement range to be resolved is fixed by the acquisition geometry (see Figure~\ref{fig:LIMA}c).

A six-level network, as used in most optical-flow networks including earlier $\LIMA$ variants, aggregates information over input patches spanning thousands of pixels (see Figure~\ref{fig:LIMA}d). For $\mathrm{SR}=2$ and a 67-pixel receptive-field decoder (as used in earlier $\LIMA$ variants~\cite{manickathan2023lightweight,Mucignat2025}), Equation~\eqref{eq:Ltot} yields $L_{\mathrm{total}} = 4{,}859$~px at level six, whereas the inter-frame displacement is of only $10$--$20$~px. Reducing the encoder to three levels and the decoder receptive field to 15~px reduces the total support at the deepest level to $L_{\mathrm{total}} \leq 203$~px and the total input-image padding from a total of $189$~px to $29$~px. The reduction in depth also lowers the deepest-layer cumulative stride from $S_6 = 64$ to $S_3 = 8$, so that the loss function is evaluated on a finer spatial \rev{grid} than in six-level variants and boundary artefacts are substantially reduced. Three levels are therefore sufficient for the displacement scales considered here, and we denote the resulting architecture $\LIMA{^{3}}$.

\subsection*{Factorial ablation of architectural design choices}
\label{sec:ablation}

To isolate the contribution of each design choice, we perform a complete factorial ablation comprising sixteen architecturally distinct variants, originating by varying network depth, decoder receptive field, correlation normalization and feature projector. \rev{Each variant is trained from scratch at full resolution under otherwise identical hyper-parameters on the synthetic image generator described in the Methods, with a base learning rate of $2\times10^{-4}$ reduced on plateau until a common floor of $3.1\times10^{-6}$ is reached (151--240~epochs), and evaluated on the same direct numerical simulation (DNS) reference data~\cite{CarlierDNS} used elsewhere in this work.} The complete results are reported in
Supplementary Table~\ref{tab:ablation_factorial_supp}.

\rev{The factorial ablation identifies two principal effects. First, the six-level configurations require an additional stabilisation of the displacement update of the decoder, $\mathcal{D}$, at the coarsest pyramid levels: with a hard or hyperbolic-tangent bound, the coarsest levels diverge or remain inactive, whereas a softsign bound allows all sixteen configurations to complete training; the three-level configurations train with the hard bound.}

% >>> CHANGED 2026-10-29: effects re-derived from the re-scored ablation CSV (re-registered GT, exact-curl reference).
\rev{Second, averaged over the eight matched pairs of each design choice on the DNS reference, correlation-volume normalisation gives the largest reduction in displacement error ($-0.013$~px, seven of eight pairs; $-0.23\times10^{-2}$ in displacement curl, six of eight pairs), followed by the 15-pixel decoder receptive field ($-0.007$~px and $-0.27\times10^{-2}$, seven of eight pairs), which gives the largest reduction in displacement-curl error, and the three-level encoder ($-0.005$~px, seven of eight pairs, and $-0.13\times10^{-2}$, five of eight pairs), while the feature projector has the smallest effect ($-0.003$~px and $-0.10\times10^{-2}$). The additional pyramid levels therefore do not improve accuracy despite more than double the end-to-end receptive field (Eq.~\ref{eq:Ltot}), confirming that they are unnecessary for the displacement scales considered here. The reduced decoder receptive field also reduces the required boundary padding, limiting the padding-related artefacts identified in the earlier architecture~\cite{Mucignat2025}. Together, these results define the configuration adopted for $\LIMA{^{3}}$: a three-level encoder, a 15-pixel decoder receptive field, a two-channel feature projector, and correlation-volume normalisation.}

\subsection*{Generalisation to public datasets used by previous learned-PIV models}

We evaluate $\LIMA{^{3}}$ on the public benchmark introduced by Cai \textit{et al.} \cite{Cai2019}, a widely used test set for learned PIV. Following \cite{lagemann2021deep}, we consider the original noise-free image set \cite{cai_piv_dataset_github}, denoted Problem~Class~1 (PC1), and the noisy variant introduced in~\cite{lagemann2021deep}, denoted Problem~Class~2 (PC2) \cite{codeocean_RAFT}.

\begin{table}[htbp]
\centering
\footnotesize
\caption{\textbf{Average endpoint error ($\overline{\epsilon}$ px, frame$^{-1}$, see Methods section) on the Problem Class 1 (PC1) and Problem Class 2 (PC2) test}.  $\LIMA{^{3,0}_{\mathrm{PC1}}}$ and $\LIMA{^{3,0}_{\mathrm{PC2}}}$ denote the level-0 architecture retrained on the public PC1 and PC2 datasets, respectively. The published entries in~(a) are the averaged endpoint errors reported in~\cite{lagemann2021deep} and~\cite{Reddy2025}.}
\begin{subtable}{\linewidth}
\centering
\caption{Problem~Class~1, per case, same subsamples as \cite{Cai2020}  ($N = 108/29/48/72/45$).}
\label{tab:cai_class1}
\begin{tabular}{l l ccccc}
\toprule
Method & Train/eval & Backstep & JHTDB-ch & DNS-turb & Cylinder & SQG \\
\midrule
\multicolumn{7}{l}{\itshape $\WIDIM$ re-run, ours} \\
$\WIDIM$ $16\times16$ & classical & 0.015 & 0.051 & 0.193 & 0.036 & 0.194 \\
$\WIDIM$ $8\times8$   & classical & 0.067 & 0.063 & 0.109 & 0.066 & 0.096 \\
\midrule

\multicolumn{7}{l}{\itshape $\LIMA{^{}}$, kinematic training} \\
% >>> CHANGED 2026-09-15: all four kinematic-training rows re-scored on the
% >>>          synpiv_vlen_v5 family carrying the warp/upsample fix:
% >>>          LIMA_L6_0_uimpffgb, LIMA_L4_0_tjhjzenz, LIMA_L4_1_vyis04um,
% >>>          LIMA_L4_2_6u5aj161 (was udrcbxpc / kcdzevik / iv3gdmd9 / 9so20qfp).
% >>> CHANGED 2026-09-29: CSCS v5b family re-score (cai_class1_aee, aee_sharp_buf32;
% >>>          interim pins L30 ep61, L31 ep67, L32 ep71, L60 ep44).
% >>> CHANGED 2026-10-21: reader half-pixel fix (the lowercase production folders were
% >>>          placed +0.5 px off); cai_class1_aee.csv rescored, aee_sharp_buf32.
% >>> CHANGED 2026-10-29: render-1.0 production LIMA rows (cai_class1_aee.csv, aee_sharp_buf32).
$\LIMA{^{6,0}}$ & cross-distr        & \rev{0.029} & \rev{0.095} & \rev{0.106} & \rev{0.053} & \rev{0.110} \\
$\LIMA{^{3,0}}$ & cross-distr        & \rev{0.023} & \rev{0.058} & \rev{0.082} & \rev{0.035} & \rev{0.081} \\
$\LIMA{^{3,1}}$ & cross-distr        & \rev{0.039} & \rev{0.097} & \rev{0.129} & \rev{0.060} & \rev{0.130} \\
$\LIMA{^{3,2}}$ & cross-distr        & \rev{0.058} & \rev{0.174} & \rev{0.233} & \rev{0.116} & \rev{0.235} \\

\midrule
\multicolumn{7}{l}{\itshape $\LIMA{^{3}}$ retrained on PC1} \\
% >>> CHANGED: whole row is now the AUGMENTED checkpoint LIMA_PC1_L4_0_8jifjreb
% >>>          (was the no-aug v5q933iz, 0.010 & 0.019 & 0.040 & 0.014 & 0.033).
% >>>          Matches the recipe of the PC2 row in Table (b): both retrained rows are
% >>>          now augmented, warm-started from their level-1 root at base_lr 1e-4.
$\LIMA{^{3,0}_{\mathrm{PC1}}}$ & in-distr  & 0.014 & 0.026 & 0.049 & 0.019 & 0.041 \\
\midrule
\multicolumn{7}{l}{\itshape \RAFT{} re-run, ours} \\
$\RAFT{_{256}}$ & in-distr  & 0.011 & 0.094 & 0.071 & 0.020 & 0.090 \\
$\RAFT{_{32}}$  & in-distr   & 0.005 & 0.012 & 0.025 & 0.013 & 0.018 \\
\midrule
\multicolumn{7}{l}{\itshape Published averaged endpoint errors} \\
$\RAFT{_{256}}$~\cite{lagemann2021deep} & in-distr & 0.016 & 0.137 & 0.093 & 0.014 & 0.117 \\
$\RAFT{_{32}}$~\cite{lagemann2021deep}  & in-distr & 0.004 & 0.011 & 0.028 & 0.018 & 0.021 \\
$\RAFT{_{256}}$~\cite{Reddy2025}        & in-distr & 0.016 & 0.165 & 0.072 & 0.014 & 0.095 \\
$\RAFT{_{32}}$~\cite{Reddy2025}         & in-distr & 0.004 & 0.013 & 0.022 & 0.011 & 0.021 \\
Twins-PIVNet~\cite{Reddy2025}           & in-distr & 0.013 & 0.092 & 0.056 & 0.012 & 0.091 \\
\bottomrule
\end{tabular}
\end{subtable}
\vspace{8pt}
\begin{subtable}{\linewidth}
\centering
\caption{Problem~Class~2, per case ($N=10$ pairs per case) from \cite{codeocean_RAFT}.}
\label{tab:cai_class2}
\begin{tabular}{l l cccccc}
\toprule
Method & Train/eval & Backstep & Cylinder & JHTDB & DNS-turb & SQG & High-disp \\

\midrule
\multicolumn{8}{l}{\itshape $\WIDIM$ re-run, ours} \\
% >>> CHANGED: DNS-turb 0.124 -> 0.123
$\WIDIM$ $16\times16$ & classical & 0.017 & 0.046 & 0.088 & 0.123 & 0.194 & 0.981 \\
$\WIDIM$ $8\times8$   & classical & 0.043 & 0.055 & 0.062 & 0.078 & 0.100 & 1.500 \\
\midrule

\multicolumn{8}{l}{\itshape $\LIMA{^{}}$, kinematic training} \\
% >>> CHANGED 2026-09-15: same v5 re-score as Table (a). LIMA^{6,0} DNS-turb
% >>>          (0.089) is the one cell that did not move.
% >>> CHANGED 2026-09-29: CSCS v5b family re-score (raft_pc2_table2_aee aee_sharp;
% >>>          raft_val256_highdisp10_aee aee_sharp_full; interim pins L30 ep61, L31 ep67,
% >>>          L32 ep71, L60 ep44).
% >>> CHANGED 2026-10-21: reader half-pixel fix (as Table (a)); raft_pc2_table2_aee.csv
% >>>          aee_sharp, raft_val256_highdisp10_aee.csv aee_sharp_full.
% >>> CHANGED 2026-10-29: render-1.0 production LIMA rows (raft_pc2_table2_aee.csv aee_sharp; High-disp raft_val256_highdisp10_aee.csv aee_sharp_full).
$\LIMA{^{6,0}}$ & cross-distr & \rev{0.026} & \rev{0.063} & \rev{0.081} & \rev{0.080} & \rev{0.118} & \rev{0.658} \\
$\LIMA{^{3,0}}$ & cross-distr & \rev{0.025} & \rev{0.040} & \rev{0.060} & \rev{0.060} & \rev{0.085} & \rev{0.731} \\
$\LIMA{^{3,1}}$ & cross-distr & \rev{0.040} & \rev{0.068} & \rev{0.101} & \rev{0.095} & \rev{0.138} & \rev{0.732} \\
$\LIMA{^{3,2}}$ & cross-distr & \rev{0.063} & \rev{0.128} & \rev{0.171} & \rev{0.166} & \rev{0.249} & \rev{0.738} \\
\midrule
\multicolumn{8}{l}{\itshape $\LIMA{^{3}}$ retrained on PC2} \\
% >>> CHANGED: whole row is now the AUGMENTED checkpoint 05ww1j8w (was zfqcvuw1,
% >>>          0.017 & 0.027 & 0.038 & 0.046 & 0.060 & 0.280)
$\LIMA{^{3,0}_{\mathrm{PC2}}}$ & in-distr   & 0.021 & 0.031 & 0.042 & 0.051 & 0.066 & 0.314 \\
\midrule
\multicolumn{8}{l}{\itshape \RAFT{} re-run, ours} \\
$\RAFT{_{256}}$ & in-distr   & 0.057 & 0.106 & 0.254 & 0.199 & 0.303 & 0.515 \\
% >>> CHANGED: SQG 0.047 -> 0.046
$\RAFT{_{32}}$  & in-distr   & 0.011 & 0.022 & 0.028 & 0.035 & 0.046 & 1.725 \\
\midrule
\multicolumn{8}{l}{\itshape Published data from \cite{lagemann2021deep}} \\
$\RAFT{_{256}}$~\cite{lagemann2021deep} & in-distr& 0.064 & 0.052 & 0.228 & 0.197 & 0.249 & 0.325 \\
$\RAFT{_{32}}$~\cite{lagemann2021deep}  & in-distr& 0.014 & 0.021 & 0.036 & 0.068 & 0.059 & 0.445 \\
\bottomrule
\end{tabular}
\end{subtable}
\end{table}

Cross-architecture comparisons on these benchmarks are confounded by differences in the training distributions. The deployed $\LIMA{^{3}}$ variants and their six-level predecessor were trained using our kinematic generator \cite{Zdyba2025}, whereas the evaluated $\RAFT{}$ checkpoints were trained on PC1 or PC2. We therefore also train $\LIMA{^{3,0}}$ separately on PC1 and PC2, denoted $\LIMA{^{3,0}_{\mathrm{PC1}}}$ and $\LIMA{^{3,0}_{\mathrm{PC2}}}$, respectively. These configurations enable direct architectural comparisons with $\RAFT{}$ under matched public training distributions and are included as controls rather than deployment configurations. A further complication is the error metric. Classical and early CNN baselines reported for this benchmark were originally evaluated using root-mean-square error \cite{Cai2019b} and were subsequently reproduced under an averaged endpoint error in \cite{lagemann2021deep,Reddy2025}. Because root-mean-square error is systematically larger than averaged endpoint error, these results are not directly comparable. We therefore report the published entries separately in Supplementary Table~\ref{tab:cai_rmse_si}, together with our own root-mean-square errors. 

The mean endpoint errors for PC1 are reported in Table~\ref{tab:cai_class1}, evaluated on the same subsample as Ref.~\cite{Cai2019}. %Among models trained with our kinematic generator, reducing the architecture from six to three levels reduces the error of the full-resolution $\LIMA{^{3,0}}$ variant in all five cases, consistent with the receptive-field analysis. 
Among models trained with our kinematic generator, the full-resolution $\LIMA{^{3,0}}$ variant has lower error than its six-level counterpart in all five cases, consistent with the receptive-field analysis. Our re-evaluation of \WIDIM{} provides a more challenging classical baseline than the published results. % >>> CHANGED 2026-10-21: reader half-pixel fix; comparisons re-derived from the rescored Table (a).
% >>> CHANGED 2026-10-29: render-1.0 production rows; Cylinder now favours LIMA^3,0 (cai_class1_aee.csv).
\rev{Against the $8\times8$-pixel interrogation window, the deployed $\LIMA{^{3,0}}$ has lower error in all five cases. Against the $16\times16$-pixel window, it has lower error in three of five cases, DNS turbulence ($0.082$ versus $0.193$~px), SQG ($0.081$ versus $0.194$~px) and Cylinder ($0.035$ versus $0.036$~px). The exceptions are the smoothest fields (Backstep, $0.023$ versus $0.015$~px, and JHTDB-channel, $0.058$ versus $0.051$~px), for which the larger correlation window performs better.} Under matched PC1 training, $\LIMA{^{3,0}_{\mathrm{PC1}}}$ outperforms $\RAFT{_{256}}$ in four of the five cases, with the largest differences for JHTDB-channel ($0.026$ versus $0.094$~px), DNS turbulence ($0.049$ versus $0.071$~px), and SQG ($0.041$ versus $0.090$~px); the exception is Backstep ($0.014$ versus $0.011$~px). $\LIMA{^{3,0}_{\mathrm{PC1}}}$ also achieves lower errors than the published Twins-PIVNet values on the three turbulent cases. By contrast, $\RAFT{_{32}}$ retains the lowest PC1 error in all five cases, consistent with its design emphasis on preserving fine spatial structures~\cite{lagemann2021deep}.

We next evaluate the PC2 test samples from Lagemann \textit{et al.}, available at \cite{codeocean_RAFT}. In addition to the five standard PC2 cases, we consider a high-displacement subset comprising ten randomly selected image pairs with peak displacements exceeding $16$~px. Because the original high-displacement test set used in Ref.~\cite{lagemann2021deep} is not publicly available, our subset cannot be used to reproduce the published benchmark directly; instead, it provides a surrogate stress test for high-displacement conditions.

The same architectural reduction improves $\LIMA{^{3,0}}$ on PC2, where it achieves lower error than its six-level counterpart in \rev{all five} standard cases, consistent with the trend observed for PC1 and with the receptive-field analysis. %Among the models trained with our kinematic generator, the three-level $\LIMA{^{3,0}}$ configuration achieves lower error than its six-level counterpart in four of the five standard PC2 cases. This matches the trend observed for PC1 and is consistent with the receptive-field analysis. 
% >>> CHANGED 2026-10-21: reader half-pixel fix; comparisons re-derived from the rescored Table (b).
\rev{Compared with $\WIDIM$, the deployed $\LIMA{^{3,0}}$ has lower error than the $8\times8$-pixel interrogation window in all five cases. It has lower error than the $16\times16$-pixel window in four of the five cases, with Backstep ($0.025$ versus $0.017$~px) the only exception. The advantage over the $16\times16$-pixel window is particularly large for the two turbulent cases: DNS turbulence ($0.060$ versus $0.123$~px) and SQG ($0.085$ versus $0.194$~px).} On the high-displacement subset, the deployed $\LIMA{^{3,0}}$ remains  more accurate than both \WIDIM{} configurations, reducing the mean endpoint error by factors of \rev{$1.3$ and $2.1$} relative to the $16\times16$- and $8\times8$-pixel windows, respectively. 

When trained directly on the benchmark distribution, $\LIMA{^{3,0}_{\mathrm{PC2}}}$ \rev{achieves lower errors than the deployed configuration in all five cases}, and outperforms $\RAFT{_{256}}$ across all evaluated PC2 cases, including the high-displacement subset. The high-resolution $\RAFT{_{32}}$ checkpoint nevertheless achieves the lowest error on the five standard PC2 cases, consistent with the PC1 results and with its design emphasis on preserving fine spatial structures~\cite{lagemann2021deep}. On the high-displacement subset, however, $\RAFT{_{32}}$ exhibits a marked increase in error, whereas $\LIMA{^{3,0}_{\mathrm{PC2}}}$ and $\RAFT{_{256}}$ remain considerably more stable. Overall, the PC2 benchmark confirms the trends observed on PC1: the three-level $ \LIMA{}$ configuration achieves lower error than its six-level counterpart, the deployed $\LIMA{^{3,0}}$ \rev{matches or improves on classical multi-pass PIV on all but the smoothest case without benchmark-specific retraining}, and training on the benchmark distribution provides further gains.

\subsection*{Cross-distribution evaluation on an independent  benchmark}
\label{sec:accuracy_synth}

We next evaluate the methods on the independently generated DNS benchmark~\cite{CarlierDNS}. The benchmark comprises 100 clean particle-image pairs with ground-truth displacement fields and was not used to train any of the evaluated models. It therefore provides a controlled test of cross-distribution sub-pixels accuracy and spatial-gradient recovery.

\begin{figure}[htbp!]
    \centering
    \includegraphics[width=1\linewidth, trim=0 450 0 0, clip]{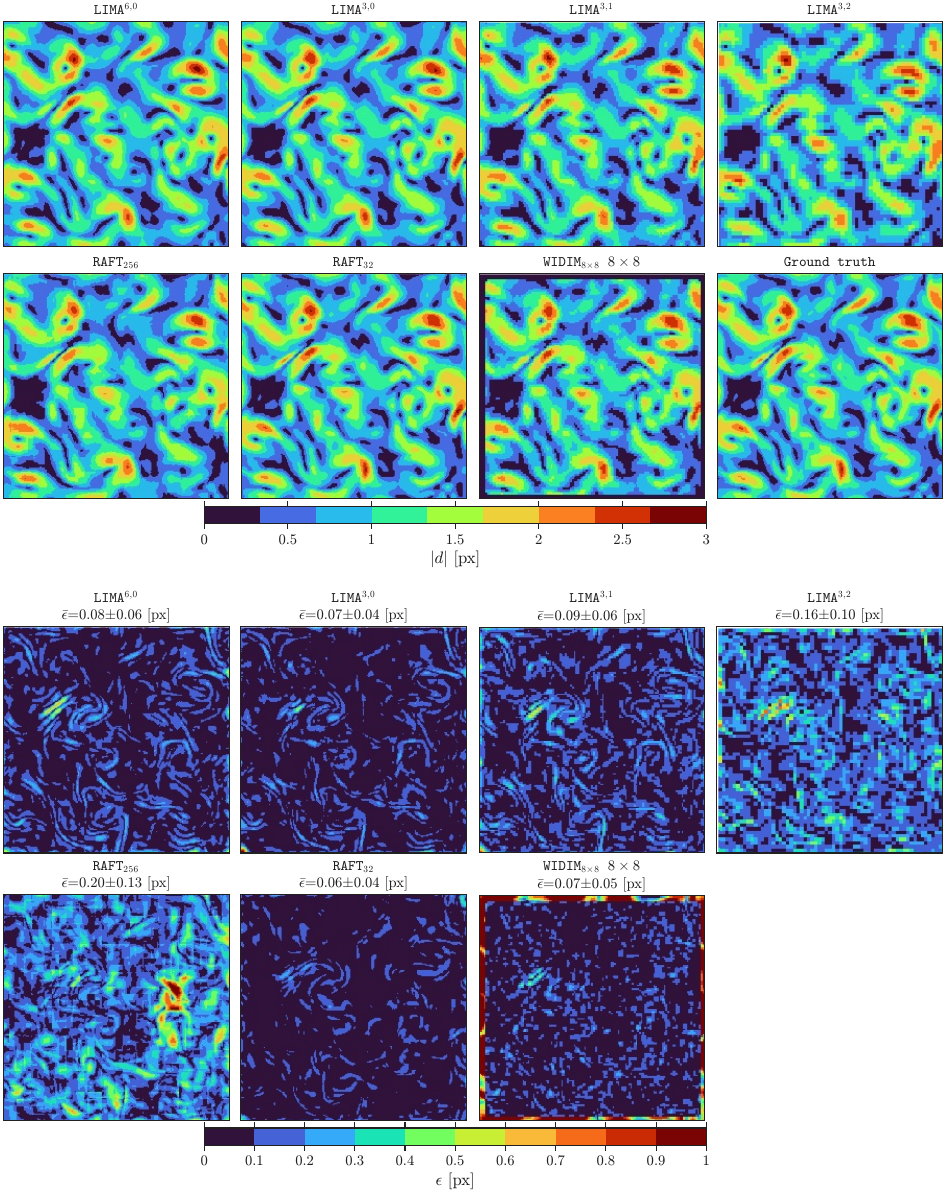}
    \includegraphics[width=1\linewidth, trim=0 295 0 239, clip]{figures/Figure2_a.pdf}
    \includegraphics[width=1\linewidth, trim=0 160 0 280, clip]{figures/Figure2_a.pdf}
    \includegraphics[width=1\linewidth, trim=0 0 0 538, clip]{figures/Figure2_a.pdf}
    \includegraphics[width=1\linewidth, trim=0 450 0 0, clip]{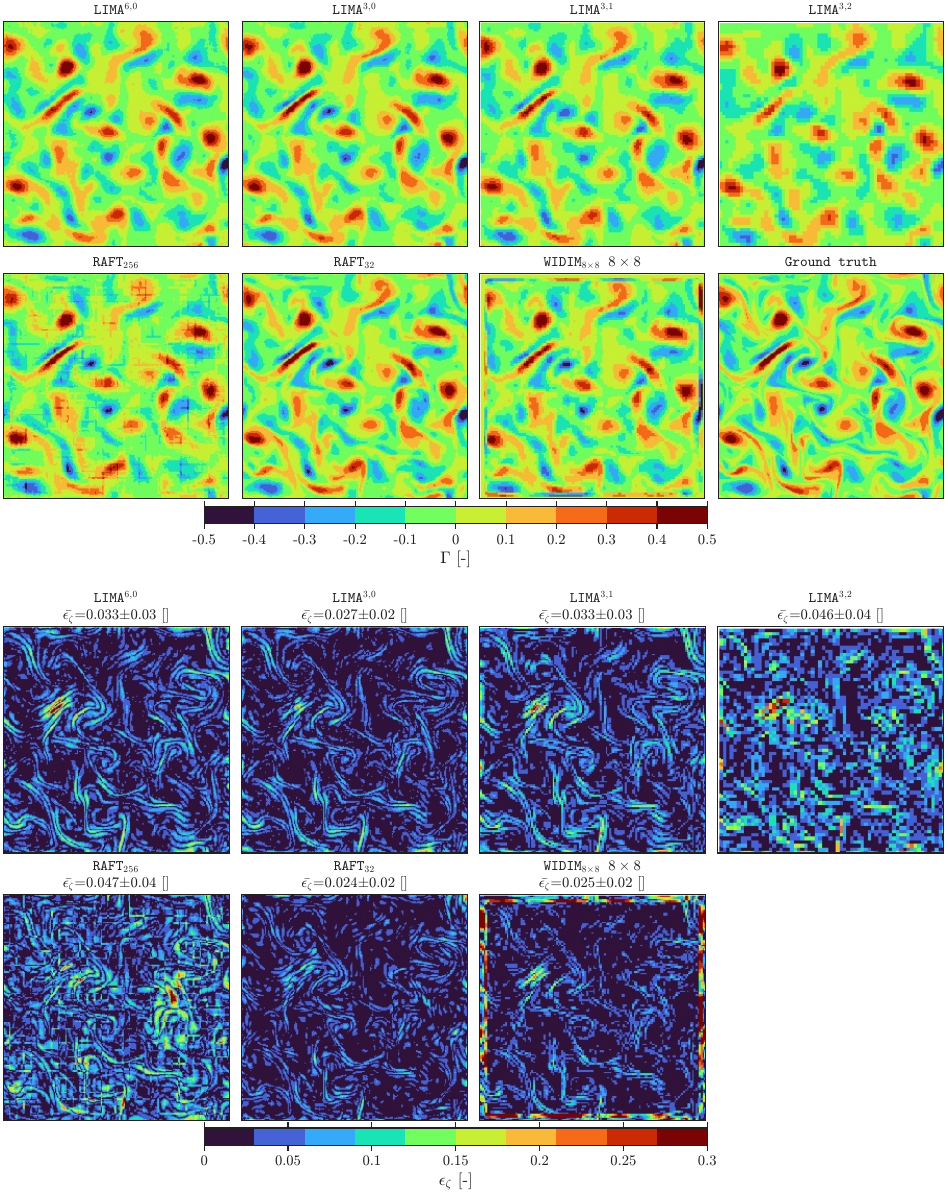}
    \includegraphics[width=1\linewidth, trim=0 295 0 239, clip]{figures/Figure2_b.pdf}
    \includegraphics[width=1\linewidth, trim=0 160 0 280, clip]{figures/Figure2_b.pdf}
    \includegraphics[width=1\linewidth, trim=0 0 0 538, clip]{figures/Figure2_b.pdf}
    \caption{\textbf{Evaluation on an independent DNS benchmark.} DNS sample No.~82 from Carlier \cite{CarlierDNS}. Rows 1 and 2: reconstructed displacement-magnitude fields for all methods and ground truth, followed by the corresponding endpoint-error maps. Rows 3 and 4: reconstructed curl fields for all methods and ground truth and corresponding curl-error maps. Methods: $\LIMA{^{6,0}}$, $\LIMA{^{3,0}}$, $\LIMA{^{3,1}}$, $\LIMA{^{3,2}}$.}
    \label{fig:DNS1}
\end{figure}

We compare the three $\LIMA{^{3}}$ variants with the six-level $\LIMA{^{6,0}}$, their immediate architectural predecessor. Complete quantitative comparisons with the two $\RAFT{}$ variants of Ref.~\cite{lagemann2021deep}, trained on PC2, and with $\WIDIM{}$ are reported over all 100 samples in Table~\ref{tab:carlier_dns}.

The reconstructed displacement magnitude for a representative DNS velocity field is shown in Figure~\ref{fig:DNS1}, together with the corresponding endpoint-error maps. \rev{$\LIMA{^{3,0}}$ shows lower endpoint errors than $\LIMA{^{6,0}}$, consistent with the conclusion from the factorial study that the three-level design improves sub-pixels accuracy relative to the six-level predecessor.} The reduced-resolution $\LIMA{^{3,2}}$ exhibits the expected loss of spatial detail.

For all samples, we additionally evaluate the curl of the reconstructed displacement field, hereafter referred to as the displacement curl. For a fixed inter-frame time, this quantity is proportional to the physical out-of-plane vorticity and therefore assesses the recovery of spatial displacement gradients. The representative displacement-curl fields and corresponding error maps are also shown in Figure~\ref{fig:DNS1}. The results show a broadly similar pattern to the displacement error: $\LIMA{^{3,0}}$ improves on the six-level predecessor, while the reduced-resolution variants incur the expected accuracy penalty.

% >>> CHANGED 2026-10-29: numbers and claims from figures/lima/turb2d_det_table.csv (render-1.0 models,
% >>>          re-registered GT, true displacement curl).
 The mean and standard deviation $\sigma$ of the displacement and displacement-curl errors over all 100 samples are reported in Table~\ref{tab:carlier_dns}. \rev{The two metrics show a similar overall performance pattern.} $\RAFT{_{32}}$ attains the lowest mean errors for displacement and displacement curl, with \rev{$0.056\pm0.033$\,px and $(2.19\pm1.87)\times10^{-2}$}, respectively. \rev{$\LIMA{^{3,0}}$ has the second-lowest displacement error, $0.066\pm0.041$\,px, lower than that of the $8\times8$ $\WIDIM{}$ configuration ($0.069\pm0.042$\,px), with a displacement-curl error of $(2.55\pm2.25)\times10^{-2}$.} \rev{Its displacement-curl error is, however, higher than that of $\WIDIM{}$ using $8\times8$ interrogation windows, $(2.30\pm2.08)\times10^{-2}$.} \rev{Against the $16\times16$-pixel interrogation window ($0.109\pm0.077$\,px and $(4.17\pm3.54)\times10^{-2}$), both $\LIMA{^{3,0}}$ and $\LIMA{^{3,1}}$ have lower displacement and displacement-curl errors.} \rev{Relative to $\LIMA{^{6,0}}$, $\LIMA{^{3,0}}$ reduces the mean displacement error from $0.080$ to $0.066$\,px and the mean displacement-curl error from $3.06\times10^{-2}$ to $2.55\times10^{-2}$, corresponding to reductions of approximately $18\%$ and $16\%$, respectively.} \rev{The half-resolution $\LIMA{^{3,1}}$ has higher displacement and displacement-curl errors than $\LIMA{^{6,0}}$ ($0.089$ versus $0.080$\,px and $3.13\times10^{-2}$ versus $3.06\times10^{-2}$)}, whereas the high-throughput $\LIMA{^{3,2}}$ shows the expected reduction in accuracy. \rev{Nevertheless, $\LIMA{^{3,2}}$ shows a lower displacement error than $\RAFT{_{256}}$ ($0.148$ versus $0.161$\,px), with a comparable displacement-curl error ($4.33\times10^{-2}$ versus $4.30\times10^{-2}$).}
\begin{table}[htbp]
\centering
\footnotesize
\caption{
\textbf{Mean displacement error ($\overline{\epsilon}$, px\,frame$^{-1}$)
and mean displacement-curl error ($\overline{\epsilon_\zeta}$, dimensionless, in units of $10^{-2}$)
on the Carlier DNS benchmark~\cite{CarlierDNS}.}
Both metrics are averaged over all 100 samples.
$\RAFT{_{32}}$ and $\RAFT{_{256}}$ are trained on PC2~\cite{lagemann2021deep};
$\WIDIM{}$ uses final interrogation windows of $8\times8$ and $16\times16$ pixels.
}
\label{tab:carlier_dns}

\begin{tabular}{lcc}
\toprule
Method &
$\overline{\epsilon}\pm\sigma_\epsilon$ [px] &
$\overline{\epsilon_\zeta}\pm\sigma_{\epsilon_\zeta}$ [$10^{-2}$] \\
\midrule

\multicolumn{3}{l}{\itshape $\LIMA{}$, kinematic training} \\
% >>> CHANGED 2026-09-15: same v5 re-score as Table~\ref{tab:cai_class1}.
% >>> CHANGED 2026-09-15: displacement to 3 dp, displacement curl rescaled to
% >>>          units of 1e-2; values themselves unchanged.
% >>> CHANGED 2026-09-29: CSCS v5b family re-score (turb2d_det, ensemble, 99 samples;
% >>>          interim pins L30 ep61, L31 ep67, L32 ep71, L60 ep44).
% >>> CHANGED 2026-10-29: render-1.0 production models; ground truth re-registered (2026-10-21);
% >>>          curl = true displacement curl vs the exact GT curl at the vector positions (score_curl.m);
% >>>          figures/lima/turb2d_det_table.csv (99 frames, as turb2d_det.m).
$\LIMA{^{6,0}}$ & \rev{$0.080\pm0.052$} & \rev{$3.06\pm2.71$} \\
$\LIMA{^{3,0}}$ & \rev{$0.066\pm0.041$} & \rev{$2.55\pm2.25$} \\
$\LIMA{^{3,1}}$ & \rev{$0.089\pm0.055$} & \rev{$3.13\pm2.68$} \\
$\LIMA{^{3,2}}$ & \rev{$0.148\pm0.089$} & \rev{$4.33\pm3.60$} \\

\midrule

\multicolumn{3}{l}{\itshape \RAFT{}, PC2, re-run, ours} \\
% >>> CHANGED 2026-10-29: RAFT and WIDIM rows re-scored for the same two reasons (GT registration, true curl).
$\RAFT{_{32}}$  & \rev{$0.056\pm0.033$} & \rev{$2.19\pm1.87$} \\
$\RAFT{_{256}}$ & \rev{$0.161\pm0.099$} & \rev{$4.30\pm3.55$} \\

\midrule

\multicolumn{3}{l}{\itshape \WIDIM{}} \\
$\WIDIM{}$ ($8\times8$)  & \rev{$0.069\pm0.042$} & \rev{$2.30\pm2.08$} \\
$\WIDIM{}$ ($16\times16$) & \rev{$0.109\pm0.077$} & \rev{$4.17\pm3.54$} \\

\bottomrule
\end{tabular}
\end{table}

\subsection*{Resolution–noise trade-off under controlled particle loss}
\label{sec:pareto}

Finer velocity fields can be recovered only at the cost of increased noise, so reconstruction quality is determined by the resulting resolution--noise trade-off. To investigate this trade-off, we use the sinusoidal-shear-flow benchmark of \cite{Mucignat2025}, in which the true vertical component is zero and the horizontal displacement is $u = A_0\sin(2\pi y/\lambda)$. For amplitudes $A_0$ of 1 and 4~pixels, we sweep the wavelength $\lambda$ from 1 to 200~pixels and quantify the resulting reconstruction by the equivalent filter length $L_{sr}$ (Methods,
  Eq.~\eqref{eq:lsr}), which characterises the spatial resolution~\cite{BernhardWieneke2017}, and the random error of the nominally zero vertical displacement component, $\Delta_V$, which quantifies noise. Each $\lambda$-sweep is summarised as a single point $(L_{sr},\Delta_V)$ in the resolution--noise plane. Synthetic particle images with a resolution of $512\times128$ pixels (the maximum size allowed by \RAFT{$_{32}$} memory requirements) are generated with random out-of-plane particle loss ranging from $0\%$ to $50\%$, mimicking a major source of correlation loss and measurement error in experimental PIV. In the case of \WIDIM{}, we report results obtained for final interrogation-window sizes of 8, 16, 24, and 32~pixels. As in the public-benchmark comparison, we separate the effects of architecture and training distribution by including $\LIMA{^{3,0}_{\mathrm{PC2}}}$, which has the \rev{same} architecture as $\LIMA{^{3,0}}$ but is trained on the PC2 distribution used for both $\RAFT{}$ variants. Note that PC2 does not include out-of-plane particle loss (Methods).

\begin{figure}[htbp!]
    \centering
    \includegraphics[width=1\linewidth]{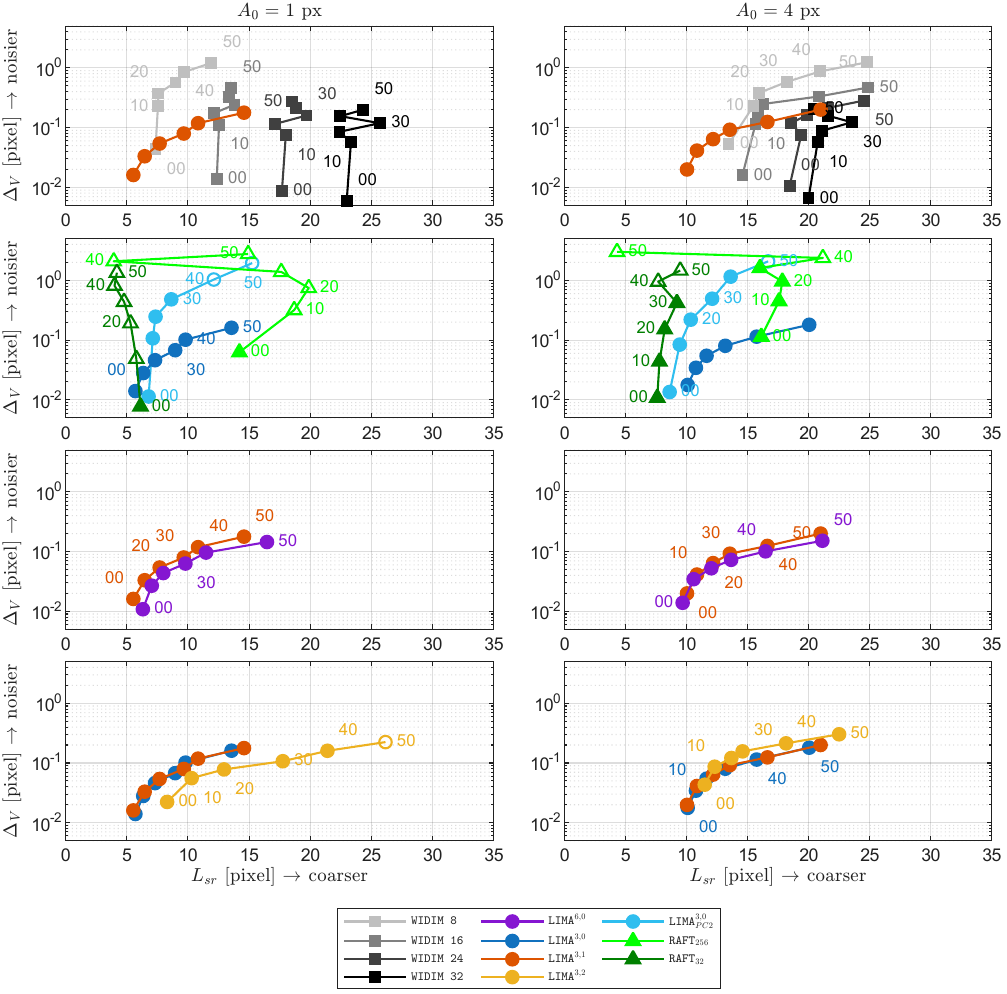}
  \caption{\textbf{Resolution--noise plane for the sinusoidal benchmark of
\cite{Mucignat2025}}. Columns: amplitude $A_0 = 1$~px (left) and $4$~px
(right). Methods: \RAFT{$_{256}$}, \RAFT{$_{32}$}, the control
\LIMA{$^{3,0}_{\mathrm{PC2}}$}, the \LIMA{$^{3}$} variants, \rev{\LIMA{$^{6,0}$}}, and the classical \WIDIM{} reference at different interrogation window sizes. Axes: equivalent filter length
$L_{sr}$ (smaller $=$ higher spatial resolution) versus random error
$\Delta_V$ of the nominally zero $v$-component (log scale), obtained by
sweeping $\lambda$ over $1$--$200$~px and fitting $A/A_0(\lambda)$; $L_{sr}$ follows~\cite{BernhardWieneke2017}. Open markers denote artefactual estimates ($L_{sr}$ below the clean $0\%$ value, or noise $>1$~px).}
    \label{fig:Amplitude_pareto}
\end{figure}
\begin{figure}[htbp]
    \centering
    \includegraphics[width=1\linewidth, trim=0 50 0 0, clip]{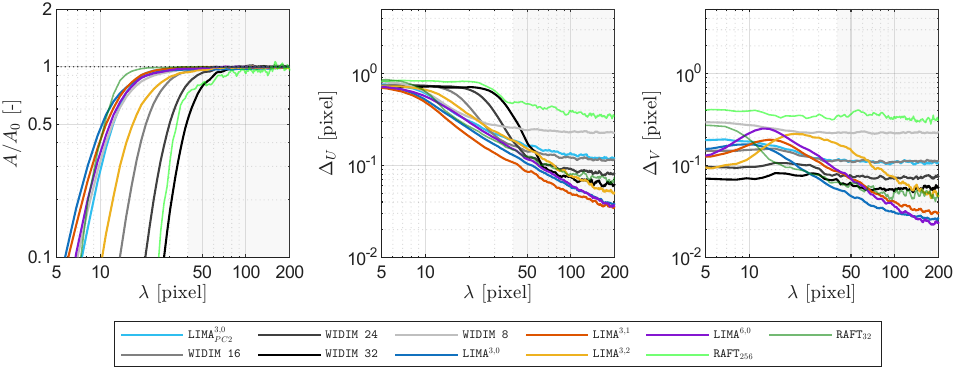}
    \includegraphics[width=1\linewidth, trim=0 52 0 0, clip]{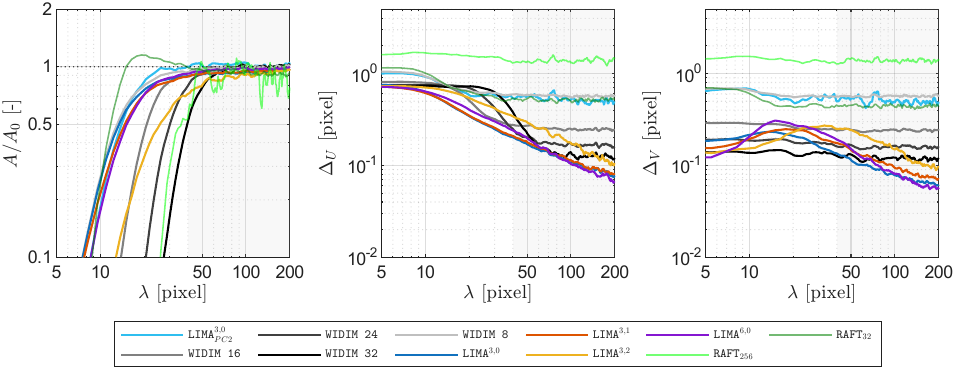}
    \includegraphics[width=1\linewidth]{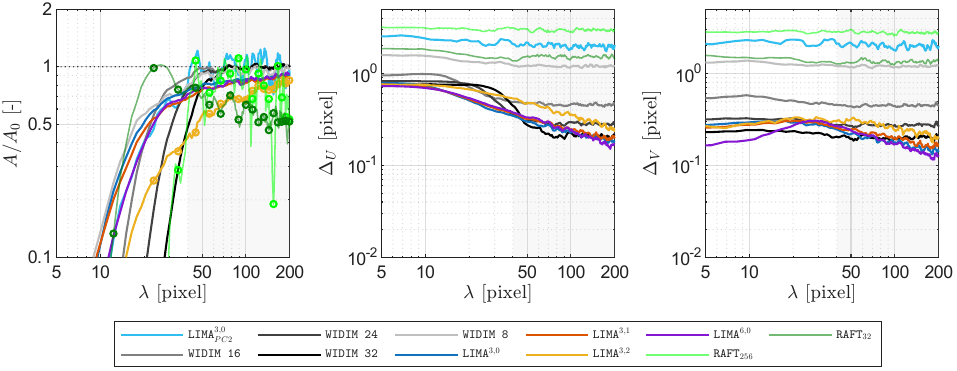}
    \caption{\textbf{Response curves for the sinusoidal benchmark at $A_0=1$~px and particle-loss levels of $10\%$ (top), $30\%$ (middle), and $50\%$ (bottom)}. Left: amplitude ratio, $A/A_0$. Middle: sine-fitting residual, $\Delta_U$. Right: random error, $\Delta_V$, of the nominally zero vertical displacement component. In all panels the abscissa is the forcing wavelength $\lambda$.}
    \label{fig:Amplitude_profiles}
\end{figure}

The resulting resolution--noise plane is shown in Figure~\ref{fig:Amplitude_pareto}, with the lower-left corner corresponding to high resolution and low noise. Trained on the same data, $\LIMA{^{3,0}_{\mathrm{PC2}}}$ and $\RAFT{_{32}}$ reach comparable noise floors at all particle-loss levels ($\Delta_V \approx 1.0$ and $0.8$~px at $A_0 = 1$ and $40\%$ loss, respectively), indicating that the training distribution strongly influences the noise level. The two architectures nevertheless show markedly different behaviour along the resolution axis. For $\LIMA{^{3,0}_{\mathrm{PC2}}}$ the equivalent filter length increases from \rev{$L_{sr} \approx 6.8$ to $15.2$~px} as particle loss increases from $0$ to $50\%$, consistent with the expected loss of spatial resolution. By contrast, $L_{sr}$ for  \RAFT{$_{32}$} decreases from $6.1$~px at $0\%$ to $\approx\!4.0$~px at $40\%$ particle loss.
This apparent increase in spatial resolution is spurious: genuine spatial resolution should degrade as noise increases, and the apparent sharpening reflects a \emph{frequency-locking} artefact. In \RAFT{$_{32}$}, random small-scale fluctuations in the noisy image pair become phase-locked to the imposed sinusoidal forcing, causing the sinusoidal fit of the reconstructed $u$ field to report spuriously high spatial resolution. The mechanism is evident in the amplitude response $A/A_0$, which exceeds unity at intermediate wavelengths (Figure~\ref{fig:Amplitude_profiles}). At $50\%$ loss it falls to about $0.5$ only at $\lambda = 200$~px, so  $L_{sr}$ is pushed to an artificially short scale. The affected $L_{sr}$ values are shown as open markers in Figure~\ref{fig:Amplitude_pareto} and are excluded from the analysis. Note that none of the $\LIMA{^{3}}$ variants show this behaviour over the investigated range of particle loss.

$\RAFT_{256}$ fails differently: consistent with its reduced resolution, it acts as a low-pass filter, retaining a coarse but faithful filter length ($L_{sr}\approx14$–$20$~px) up to $\approx20\%$ particle loss, increasing as particle loss rises. Beyond this point, the recovered field progressively collapses toward zero, with $A/A_0$ falling from $0.96$ to $0.62$ by $50\%$ loss and large residuals \rev{(\cref{fig:Amplitude_profiles})} making the calculated $L_{sr}$  unreliable. Both variants therefore break down under heavy particle loss, but by opposite mechanisms: over-amplification and frequency locking for $\RAFT_{32}$, and field collapse for $\RAFT_{256}$.

Excluding the artefactual points, the trajectories in Figure~\ref{fig:Amplitude_pareto} reveal a clear pattern. % >>> CHANGED 2026-10-29: render-1.0 production rows; figures/lima/sinewave_lsr_table.csv (Lsr_AF, noise).
\rev{In the absence of particle loss, \RAFT{$_{32}$} provides the highest spatial resolution at $A_0=4$~px ($L_{sr}=7.6$~px), whereas at $A_0=1$~px $\LIMA{^{3,1}}$ and $\LIMA{^{3,0}}$ resolve slightly finer scales ($5.5$ and $5.7$ versus $6.1$~px); however, the behaviour of \RAFT{$_{32}$} changes markedly under particle loss.} As loss increases, the three-level $\LIMA{^{3,0}}$ and $\LIMA{^{3,1}}$ variants occupy the favourable lower-left region of the resolution--noise plane, showing progressively coarser spatial resolution with increasing $L_{sr}$. By contrast, $\RAFT{_{32}}$ exhibits an apparent increase in spatial resolution with decreasing $L_{sr}$, reflecting the frequency-locking artefact indicated by the open markers. The $\WIDIM$ family makes the resolution--noise trade-off explicit: larger interrogation windows reduce noise at the expense of resolution but do not reach the performance of \rev{$\LIMA{^{3,0}}$ and $\LIMA{^{3,1}}$}. \rev{$\LIMA{^{3,1}}$ and $\LIMA{^{3,0}}$ also resolve finer scales than the deeper $\LIMA{^{6,0}}$, with a smaller $L_{sr}$ at eight and ten of the twelve sweep points, respectively (including all six at $A_0=1$~px), at a moderately higher noise level (up to $49\%$ and $29\%$ higher $\Delta_V$, respectively), further supporting the receptive-field analysis.}

The choice of the training distribution strongly affects the attainable noise floor. The $\LIMA{^{3,0}_{\mathrm{PC2}}}$ noise floor is about an order of magnitude higher than that of $\LIMA{^{3,0}}$ trained on our particle-loss-aware generator (for example, \rev{$\Delta_V = 1.95$ versus $0.16$~px} at $A_0=1$ and $50\%$ loss). The comparison shows that the training distribution strongly influences the noise floor, whereas the architecture governs how the measured resolution responds to particle loss. Trained on the same PC2 data as $\RAFT$, $\LIMA{^{3,0}_{\mathrm{PC2}}}$ inherits the elevated noise floor but still coarsens as expected, with $L_{sr}$ increasing with particle loss, whereas \RAFT{$_{32}$} exhibits frequency locking and a corresponding decrease in $L_{sr}$.

\subsection*{Experimental validation}
\label{sec:experimental}
We evaluate the deployed $\LIMA{^{3}}$ on sixteen experimental flow configurations spanning separated flows over bluff bodies \cite{manickathan2023lightweight}, flow over hills \cite{lagemann2021deep}, a cable-car wind-tunnel model \cite{Barbe2026}, cylinder wakes \cite{manickathan2022kinematic}, urban canopy flow, the flow past a robotic fish \cite{Schwab2022}, and nine cases from PIV Challenges 1--4~\cite{PIVChallenge}. Four representative cases are shown in the main text (Figure~\ref{fig:Exp1}), while the remaining twelve are reported in Supplementary Figures~\ref{fig:Exp2_supp}--\ref{fig:Exp4_supp}. 
For all cases, we also report results for \RAFT{$_{256}$} and  \WIDIM{}.

\begin{figure}[htbp]
    \centering
    \includegraphics[width=0.96\linewidth, trim=0 25 0 0, clip]{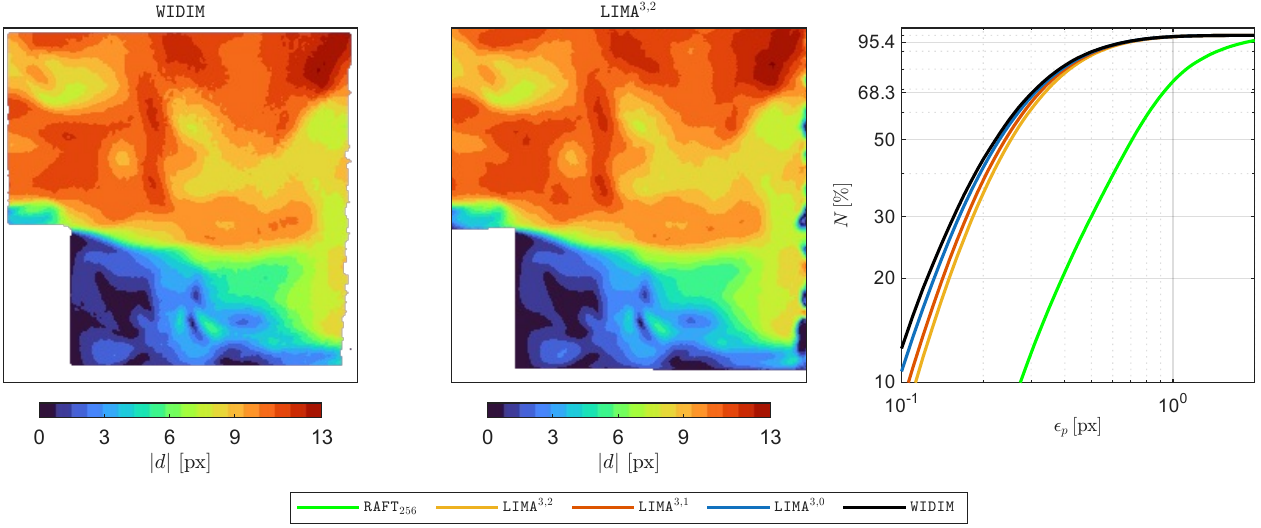}
    \includegraphics[width=0.96\linewidth, trim=0 25 0 10, clip]{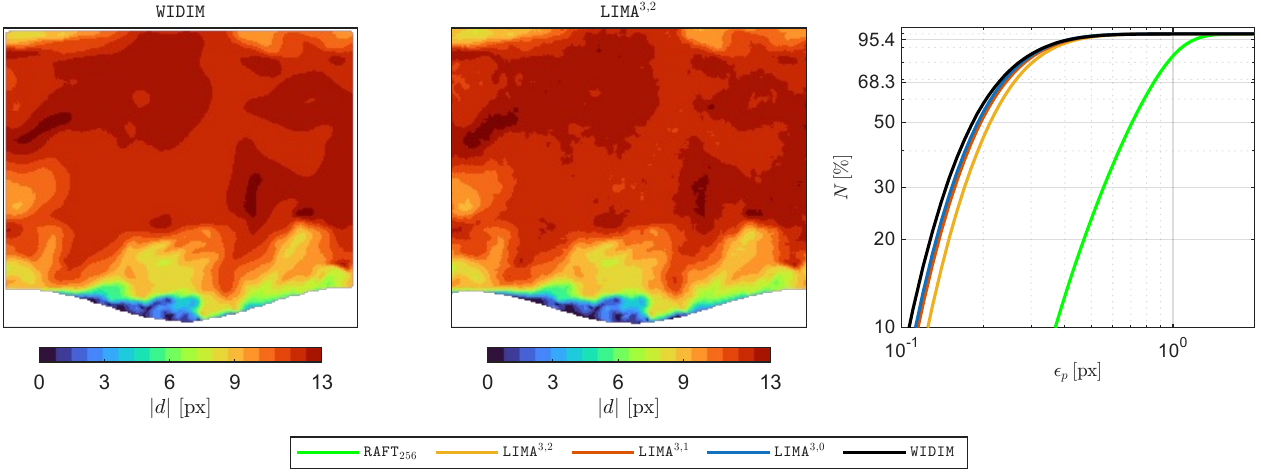}
    \includegraphics[width=0.96\linewidth, trim=0 25 0 10, clip]{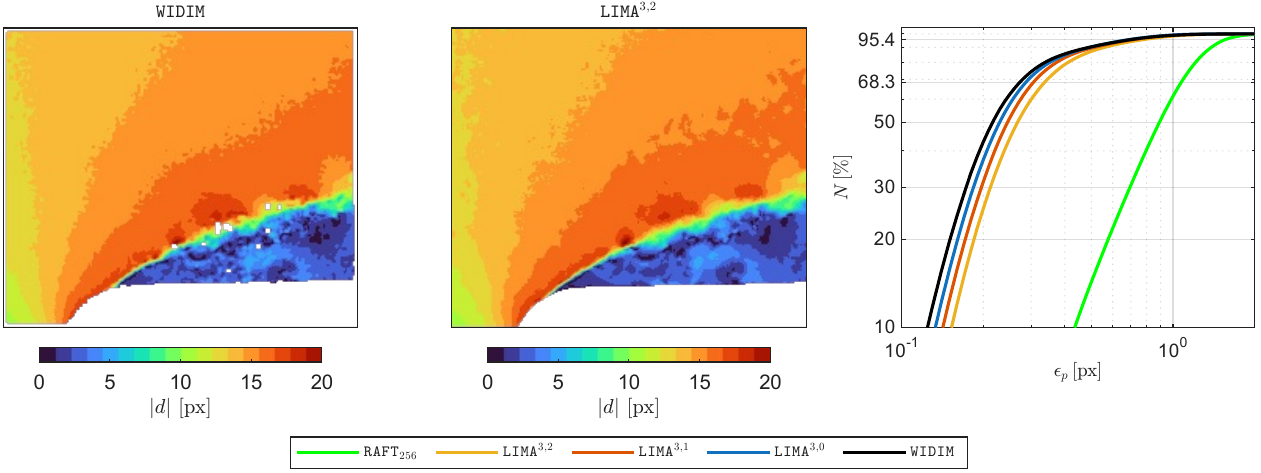}
    \includegraphics[width=0.96\linewidth, trim=0 0 0 10, clip]{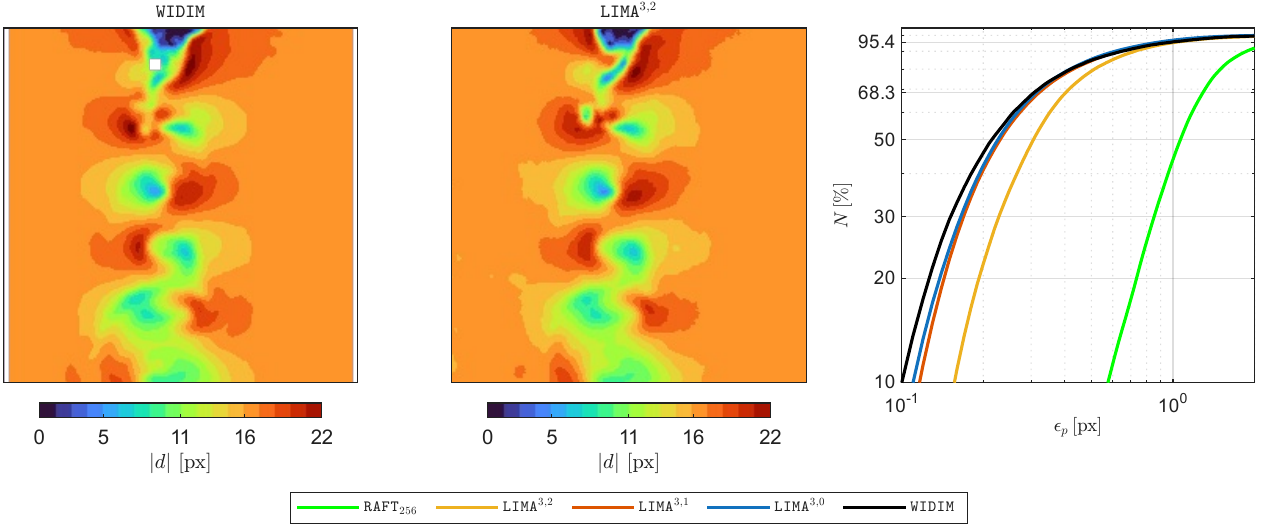}
    \caption{\textbf{Experimental validation on four representative flow configurations.} By row: bluff body in a turbulent boundary layer, flow over hills \cite{lagemann2021deep}, cable-car model, and cylinder wake. Left: displacement magnitude reconstructed by $\WIDIM$. Middle: displacement magnitude reconstructed by $\LIMA{^{3,2}}$. Right: disparity-error, $\epsilon_p$~\cite{Sciacchitano2013}, cumulative plots for $\WIDIM$,  the three $\LIMA$ variants, and \RAFT{$_{256}$}.}
    \label{fig:Exp1}
\end{figure}
Because ground-truth velocity fields are unavailable for the experimental cases, reconstruction quality is assessed using two independent approaches. The first is the disparity error, $\epsilon_p$~\cite{Sciacchitano2013}, computed by advecting particles from the first image forward and from the second image backwards to an intermediate time using the reconstructed displacement field, and measuring the residual offset between the resulting particle positions. The second comprises three matching-free, image-based indicators computed from the symmetrically warped image pair: the correlation peak ratio (PR), the primary correlation coefficient, $c$, and the residual displacement, $|d_{\mathrm{res}}|$. High-quality reconstructions are characterised by high PR and $c$, together with a sharply peaked $|d_{\mathrm{res}}|$ distribution centred near zero. 

Across the four cases in Figure~\ref{fig:Exp1}, the disparity-error distributions of $\LIMA{^{3,0}}$ and
$\LIMA{^{3,1}}$  are similar to those of $\WIDIM$, whereas the throughput-optimised $\LIMA{^{3,2}}$ shows slightly larger errors, consistent with its coarser output resolution. By contrast, $\RAFT{_{256}}$ yields substantially larger $\epsilon_p$, indicating poorer consistency with the particle motion. 
\begin{figure}[htbp]
    \centering
    \includegraphics[width=0.96\linewidth, trim=0 0 0 0, clip]{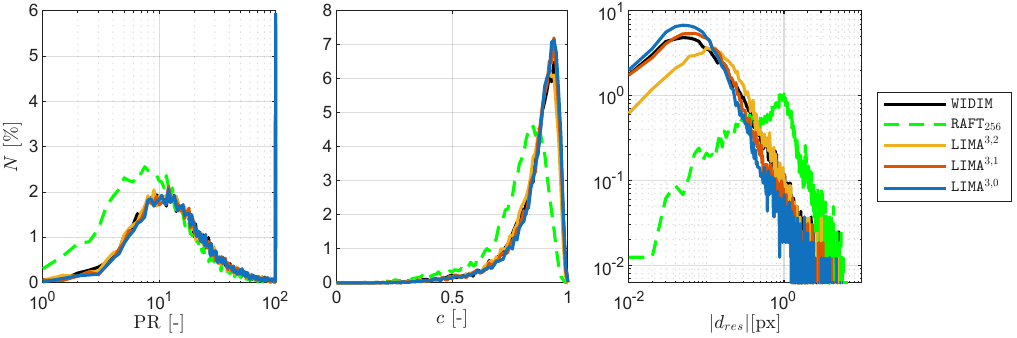}
    \caption{\textbf{Matching-free quality indicators for the cylinder-wake case, computed on the symmetrically warped image pair.} Left: correlation peak ratio (PR). Middle: primary correlation coefficient, $c$. Right: residual displacement, $|d_{\mathrm{res}}|$. The $\LIMA$ variants and $\WIDIM$ are consistent across all three indicators, while \RAFT{$_{256}$} yields lower PR and $c$ values and a broader residual-displacement distribution.}
    \label{fig:ExpPR}
\end{figure}
Figure~\ref{fig:ExpPR} reports the matching-free indicators for the cylinder wake and shows that the $\LIMA{^{3}}$ variants exhibit PR, $c$, and $|d_{\mathrm{res}}|$ distributions closely matching those of $\WIDIM$. $\LIMA{^{3,2}}$ retains comparable correlation quality, with only marginally broader $|d_{\mathrm{res}}|$ distribution. By contrast, \RAFT{$_{256}$} shows poorer \rev{matching-free} indicators, consistent with poorer agreement between the symmetrically warped particle images. The same disparity-error ranking is observed across the twelve additional experimental configurations reported in Supplementary Figures~\ref{fig:Exp2_supp}--\ref{fig:Exp4_supp}, including PIV Challenge cases spanning large-displacement boundary-layer flows, multi-camera configurations, free-surface flows, and biological flows.

\subsection*{Memory footprint, throughput, and edge deployment}
\label{sec:throughput}

\begin{figure}[htbp!]
    \centering
    \includegraphics[width=0.75\linewidth, trim=0 0 0 10.5, clip]{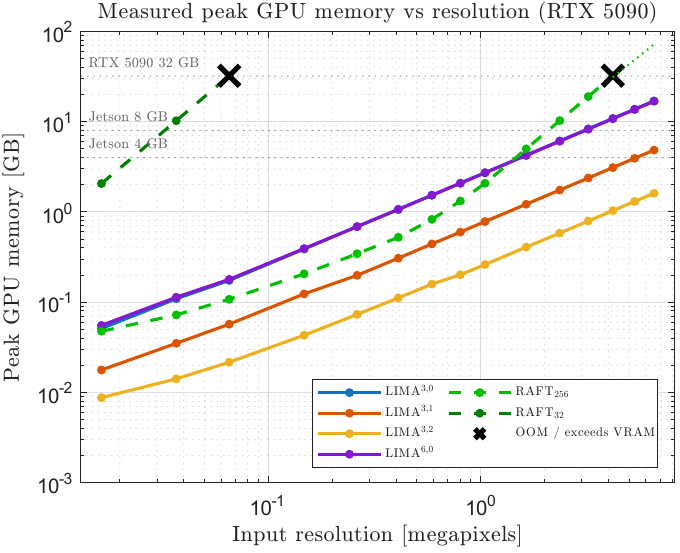}
    \caption{\textbf{Peak GPU memory versus input resolution for a single image pair on an  NVIDIA RTX~5090.}  Crosses indicate the largest input resolution accommodated within 32~GB of GPU memory: $4.2$~MPx for $\RAFT{_{256}}$ and $256\times256$ pixels for $\RAFT{_{32}}$. The figure also reports the available memory on NVIDIA Jetson for edge deployment. \LIMA{$^{3,0}$} and \LIMA{$^{6,0}$} are superimposed as the cost for inference is dominated by the cost of Layer 0.}
    \label{fig:mem_envelope}
\end{figure}

For real-time deployment alongside experiments, memory consumption is a primary constraint because the estimator must operate within the capacity of the available hardware. In $\RAFT{}$, memory is dominated by the all-pairs correlation volume, which stores the feature similarity---the inner product of the learned feature descriptors---for every pair of locations across the two feature maps; because every location in one feature map is coupled to every location in the other, memory usage grows quadratically with input image dimensions. By contrast, the $\LIMA{^{3}}$ stores only feature maps and a bounded local correlation volume, enabling linear memory scaling. Figure~\ref{fig:mem_envelope} shows peak GPU memory as a function of input resolution. All $\LIMA{^{3}}$ variants scale linearly  (log--log slope $\approx 1$) and remain within a few gigabytes across the full range, up to 6.5~megapixels. The memory footprint of $\RAFT{}$ scales quadratically (log--log slope $\approx 2$), exhausting $32$~GB of device memory at $4.2$~megapixels for $\RAFT{_{256}}$ and already at $256\times256$ pixels for $\RAFT{_{32}}$. Within the range where both architectures fit into GPU memory, the difference exceeds two orders of magnitude; at $256\times256$ pixels, for example, $\LIMA{^{3,1}}$ requires only $0.06$~GB. The quadratic scaling is an intrinsic consequence of $\RAFT$'s all-pairs correlation architecture; implementation choices can reduce the memory constant or trade memory for additional computation, but they cannot change this asymptotic dependence.

The linear memory scaling of $\LIMA{^{3}}$ enables deployment on hardware with limited memory where heavier architectures cannot be accommodated. Compared with typical embedded platforms, such as NVIDIA Jetson devices with 4 or 8~GB of memory (Figure~\ref{fig:mem_envelope}), $\LIMA{^{3,1}}$ and $\LIMA{^{3,2}}$ process megapixel image pairs within a single-digit-gigabyte memory budget, whereas the all-pairs correlation architectures of comparable accuracy exceed these memory budgets. At $640\times480$ pixels, typical of high-speed PIV recording, the lightweight $\LIMA{^{3}}$ variants require only tens of megabytes, leaving substantial memory available for image acquisition and buffering. 

The favourable memory scaling of $\LIMA{^{3}}$ is accompanied by higher inference throughput. Table~\ref{tab:speed}(a) reports the corresponding inference timings measured as the median of $30$ runs following $10$ warm-up iterations.  On the same GPU and PyTorch implementation, $\LIMA{^{3,0}}$ is faster than $\RAFT{_{256}}$ at all tested image resolutions, ranging from a 2-fold speedup at $640\times480$ pixels to 1.5-fold speedup at $1280\times1024$ pixels. The lightweight $\LIMA{^{3,2}}$ achieves the highest throughput, reaching 6-fold and 12-fold speedups at these resolutions, respectively, consistent with the increasing cost of the all-pairs correlation mechanism at higher resolution. The recent transformer-based Twins-PIVNet ($11.6$~M parameters ~\cite{Reddy2025}) is reported to achieve throughput comparable to $\RAFT{_{256}}$ at $256\times256$ pixels. Although exact inference comparisons depend on hardware and implementation details, the overall trend is consistent: $\LIMA{}$ provides substantial throughput gains at megapixel resolutions compared with larger optical-flow architectures.

\begin{table}[htbp]
    \caption{\textbf{Median inference time per image pair on a single NVIDIA RTX~5090 GPU (median $\pm$ s.d. over 30 timed iterations following 10 warm-up iterations, batch size 1)}.
    (a)~PyTorch implementation, $\RAFT{_{256}}$ against the $\LIMA{^{3}}$.
    (b)~TensorRT-compiled $\LIMA$ \rev{variants.}}
    \label{tab:speed}
    \centering
    {\fontsize{10}{10}\selectfont
    \begin{subtable}[t]{\linewidth}
    \centering
    \caption{PyTorch implementation}
    \label{tab:speed_eager}
    \begin{tabular}{lcccc}
    \toprule
    Size & $\RAFT{_{256}}$ & $\LIMA{^{3,0}}$ & $\LIMA{^{3,1}}$ & $\LIMA{^{3,2}}$ \\
    $[\mathrm{px}]$ & $[\mathrm{ms}]$ & $[\mathrm{ms}]$ & $[\mathrm{ms}]$ & $[\mathrm{ms}]$ \\
    \midrule
    $640 \times 480$    & $36.4 \pm 0.9$ & $18.1 \pm 0.4$  & $8.8 \pm 0.7$  & $5.9 \pm 0.5$  \\
    $1024 \times 1024$  & $74.7 \pm 0.5$ & $48.4 \pm 0.4$  & $15.4 \pm 0.4$ & $7.1 \pm 0.4$  \\
    $1280 \times 1024$  & $89.5 \pm 2.6$ & $59.1 \pm 0.2$  & $17.9 \pm 0.3$ & $7.4 \pm 0.4$  \\
    $2048 \times 2048$  & OOM            & $185.3 \pm 0.7$ & $55.8 \pm 0.6$ & $18.4 \pm 0.2$ \\
    $2560 \times 2160$  & OOM            & $244.8 \pm 1.2$ & $73.1 \pm 0.2$ & $23.6 \pm 0.2$ \\
    \bottomrule
    \end{tabular}
    \end{subtable}

    \vspace{1em}

    \begin{subtable}[t]{\linewidth}
    \centering
    \caption{TensorRT-compiled}
    \label{tab:speed_trt}
    \begin{tabular}{lcccc}
    \toprule
    Size & $\LIMA{^{6,0}}$ & $\LIMA{^{3,0}}$ & $\LIMA{^{3,1}}$ & $\LIMA{^{3,2}}$ \\
    $[\mathrm{px}]$ & $[\mathrm{ms}]$ & $[\mathrm{ms}]$ & $[\mathrm{ms}]$ & $[\mathrm{ms}]$ \\
    \midrule
    $640 \times 480$    & $15.6 \pm 0.2$  & $9.3 \pm 0.1$   & $3.2 \pm 0.05$  & $1.8 \pm 0.05$  \\
    $1024 \times 1024$  & $35.4 \pm 0.3$  & $29.1 \pm 0.05$  & $8.4 \pm 0.05$  & $3.8 \pm 0.05$  \\
    $1280 \times 1024$  & $42.0 \pm 0.2$  & $36.1 \pm 0.1$  & $10.2 \pm 0.05$ & $4.4 \pm 0.05$  \\
    $2048 \times 2048$  & $126.5 \pm 0.2$ & $116.0 \pm 0.1$ & $33.0 \pm 0.1$ & $13.9 \pm 0.1$ \\
    $2560 \times 2160$  & $166.9 \pm 0.2$ & $153.7 \pm 0.1$ & $44.4 \pm 0.1$ & $18.7 \pm 0.1$ \\
    \bottomrule
    \end{tabular}
    \end{subtable}
    }
\end{table}

Further gains are achieved after TensorRT \rev{compilation (Table~\ref{tab:speed}(b))}. With full resolution output, $\LIMA{^{3,0}}$ is $10$--$20\%$ faster than $\LIMA{^{6,0}}$, indicating that the reconstruction stages contribute substantially to the computational cost at higher resolutions. Reducing output resolution yields substantially larger gains: at $1024\times1024$ pixels, $\LIMA{^{3,1}}$ reaches $119$~fps ($4.2$-fold faster than $\LIMA{^{6,0}}$), whereas $\LIMA{^{3,2}}$ reaches $263$~fps ($9.3$-fold faster). At $640\times480$ pixels, $\LIMA{^{3,2}}$ exceeds $550$~fps for single image pairs and $900$~fps with a batch size of 8. 

We also compared $\LIMA{^{3,2}}$  with the GPU implementation of $\WIDIM$. Because the internal processing stages of the commercial code (DaVis) are not exposed, its computational time cannot be isolated. We therefore report an end-to-end benchmark in which both methods process the same input image files and generate the same output file format. The reported benchmark is conservative with respect to $\LIMA{^{3,2}}$: profiling indicates that approximately 36\% of the measured wall-clock time is attributable to file I/O  rather than to network inference.  For the Cylinder dataset ($2016 \times 1008$ px), $\LIMA{^{3,2}}$ processed 700 image pairs in 58 s (12.1 image pairs $s^{-1}$), whereas DaVis required 14 min (0.83 image pairs $s^{-1}$). For the Cube dataset ($2016 \times 2016$~px), $\LIMA{^{3,2}}$ processed 100 image pairs in 9 s (11.1 image pairs $s^{-1}$), compared with 2 min 50 s for DaVis (0.59 image pairs $s^{-1}$). Despite including file-read and export overheads, $\LIMA{^{3,2}}$ is 14.5--18.9-fold faster. 

$\LIMA{^{3,2}}$ also runs directly on embedded hardware. Compiled  on an NVIDIA Jetson AGX Orin 32~GB, it sustains $99$~image pairs per second at $640\times480$~pixels in the $50$~W power mode ($141$~fps in unconstrained MAXN mode), tens of image pairs per second at \rev{megapixel} resolutions, and $6.3$~fps at $2560\times2160$~pixels (Table~\ref{tab:jetson}). The same estimator therefore runs concurrently with acquisition on both workstation and edge hardware, rather than only during offline post-processing.

\begin{table}[htbp]
    \caption{\textbf{Median end-to-end inference time    per image pair for \LIMA{$^{3,2}$} on an NVIDIA Jetson AGX Orin}. Median inference time  per image pair for the three power modes.}
    \label{tab:jetson}
    \centering
    {\fontsize{10}{10}\selectfont
    \begin{tabular}{lccc}
    \toprule
    Size & 30~W & 50~W & MAXN \\
    $[\mathrm{px}]$ & $[\mathrm{ms}]$ & $[\mathrm{ms}]$ & $[\mathrm{ms}]$  \\
    \midrule
    $640 \times 480$    & 21.3  & 10.1  & 7.1   \\
    $1024 \times 1024$  & 66.7  & 31.5  & 23.0   \\
    $1280 \times 1024$  & 82.7  & 39.7  & 28.3   \\
    $2048 \times 2048$  & 258.9 & 119.7 & 89.5   \\
    $2560 \times 2160$  & 340.6 & 157.8 & 119.4  \\
    \bottomrule
    \end{tabular}
    }
\end{table}
%%%%%%%%%%%%%%%%%%%%%%%%%%%%%%%%%%%%%%%%%%%%%%%%%%%%%%%%%%%%%%%%%%%%%
%% DISCUSSION                                                       %
%% New section, ~400 words. Required by NatComm structure.          %
%%%%%%%%%%%%%%%%%%%%%%%%%%%%%%%%%%%%%%%%%%%%%%%%%%%%%%%%%%%%%%%%%%

\section*{Discussion}
\label{sec:discussion}

The receptive-field analysis provides a transferable design principle for optical-flow architectures. When the relevant displacement scale is limited by the acquisition geometry, as in PIV, Equation~\eqref{eq:Ltot} provides a spatial-support budget for selecting the encoder depth, correlation search span, and decoder support from the relevant physical scale, rather than adopting them from generic computer-vision architectures. The principle is not specific to PIV: refractive displacements in background-oriented Schlieren (BOS)~\cite{Raffel2015} are often limited to a few pixels, smaller than typical PIV motion, and therefore require less spatial support. The BOS depth sweep is consistent with this prediction, with image SNR replacing particle loss as the controlled degradation (Supplementary Figure~\ref{fig:bos_pareto_supp}). More generally, each imaging modality imposes a characteristic displacement bound and, consequently, a corresponding useful spatial support. The same accounting extends to attention-based architectures when attention is restricted to a bounded window, with the window width replacing the correlation search range. For global matching, however, the physical displacement bound does not by itself define the extent of the matching operation; the relevant question is whether correspondences beyond the admissible displacement range improve or degrade the reconstruction.

%%%%%

For \LIMA{}, three pyramid levels provide sufficient correlation search range to resolve the relevant displacements of $10$--$20$\,px. Further increasing the depth enlarges the end-to-end spatial support, cumulative stride, boundary padding, and computational cost without providing a consistent reconstruction benefit. Beyond this requirement, greater depth compromises training stability, spatial resolution under particle loss, and inference efficiency. Repeated downsampling and convolution progressively coarsen the particle-image representation, reducing the local tracer structure available for reliable matching. The resulting correlation volumes may therefore contain poorly resolved peaks, leading to artefactual displacements in the decoder output. These correlations also enter the optimisation through the multi-level loss. Correlation-volume normalisation maps the correlations at each level to a controlled dynamic range before decoding, reducing the sensitivity of the optimisation to level-dependent variations in correlation magnitude. \rev{In the factorial ablation, this gives the largest reduction in displacement error among the four design choices, whereas stable training of the six-level architecture requires a softsign bound on the displacement update of the decoder, $\mathcal{D}$, at the coarsest levels.}

%%%%%

Particle loss is a major source of error in PIV because it reduces the correspondence between frames. Unlike feature aggregation, particle loss degrades the input images and therefore affects all correlation-based methods. Nevertheless, the architectures respond differently to this degradation. $\RAFT{_{32}}$ exhibits a spurious increase in spatial resolution: its reconstructed amplitude becomes locked to the forcing frequency and can overshoot the imposed amplitude, causing the equivalent filter length to decrease nonphysically as particle loss deteriorates image quality. $\RAFT{_{256}}$ fails differently:  beyond moderate loss, the estimated \rev{field collapses} toward zero. The PC2-trained $\LIMA{^{3,0}_{\mathrm{PC2}}}$ shares the architecture with $\LIMA{^{3,0}}$ but is trained on the same Problem Class~2 distribution as $\RAFT{}$, which does not include out-of-plane particle loss. Its noise floor is similar to $\RAFT{_{32}}$, yet its equivalent filter length still increases monotonically as particle information is removed. Thus, the training distribution strongly influences the attainable noise floor, whereas the response of spatial resolution to information loss is primarily architecture dependent.

%%%%%

Although each correlation lookup is local, iterative refinement couples the estimate across the field, giving  $\RAFT{}$  an effective receptive field that extends well beyond the physical displacement range. Restricting $\RAFT{}$ inference to tiled windows progressively suppresses the frequency-locking artefact, indicating that it depends on the spatial extent of the effective receptive field (Supplementary Note and Supplementary Fig.~\ref{fig:S_windowing}). Under particle loss, this long-range coupling can reinforce spurious structure at the forcing wavelength, leading to the over-unity amplitude response. This failure mode is absent when the receptive field is bounded: neither the compact coarse-to-fine $\LIMA{^{3}}$ nor the finite interrogation window of $\WIDIM{}$ exhibits frequency locking. $\RAFT{_{256}}$ avoids the artefact through greater down-sampling, but the estimated field collapses under heavy loss. Training on particle-loss-aware data further lowers the attainable noise floor and improves robustness to information loss, but is not required for the physically expected coarsening response, as demonstrated by $\LIMA{^{3,0}_{\mathrm{PC2}}}$. 

%%%%%

On experimental images, the deployed $\LIMA{^{3}}$ variants show particle-disparity distributions similar to those of $\WIDIM{}$, while the cylinder-wake matching-free indicators show comparable correlation quality and residual-displacement distributions. By contrast, $\RAFT{_{256}}$ exhibits substantially poorer consistency with the observed particle motion. \rev{The finer effective spatial resolution of the $\LIMA{^{3}}$ variants also improves the reconstruction of flow features close to a wall: in the cable-car case the separated shear layer is recovered as a narrower displacement-curl ridge (Supplementary Fig.~\ref{fig:separation_supp}).} The shallower $\LIMA{^{3}}$ variants are lightweight, with only $589{,}000$ parameters, and memory-efficient, supporting deployment alongside experiments and on edge hardware. Replacing all-pairs correlation with a compact pyramidal feature extraction and correlation architecture enables linear memory scaling and efficient computation. The receptive-field analysis provides a principled basis for the architecture: rather than compressing an over-parameterised network or fixing the pyramid depth by convention, it determines the useful depth from the bounded displacement range of the measurement.

%%%%%%

For practical deployment, the three \LIMA{$^{3}$} variants span the accuracy--throughput--resolution trade-off, allowing the estimator to be matched to the experiment rather than the experiment to the estimator. \LIMA{$^{3,0}$} offers the highest accuracy, \LIMA{$^{3,1}$} a favourable accuracy--throughput compromise, and \LIMA{$^{3,2}$} the highest throughput for megapixel-scale images or high-frame-rate sequences, at the cost of slightly coarser spatial resolution. In end-to-end processing benchmarks, including file reading and export, \LIMA{$^{3,2}$} sustained a 14.5--18.9-fold throughput advantage over a commercial GPU implementation of \WIDIM{}. With throughput exceeding 900 fps at $640\times480$~pixels on an RTX5090, \LIMA{$^{3,2}$} can support online processing in high-speed wind-tunnel experiments, industrial test rigs, and other settings where conventional methods cannot keep pace with image acquisition. \rev{This performance} also extends to embedded hardware: \LIMA{$^{3,2}$} sustains $99$~image pairs per second at $640\times480$~pixels on an NVIDIA Jetson AGX Orin. Together, the linear memory scaling and real-time throughput support operation alongside image acquisition on both workstation and embedded hardware, rather than only during post-processing.

%%%%%%

The analysis is derived for single-camera 2D PIV. Stereo and tomographic configurations have different correlation geometries and would require the spatial-support bound to be re-derived accordingly. The resolution--noise sweep is formulated for periodic forcing, and aperiodic flows would require an alternative to the sinusoidal fit used here. Deeper networks are not intrinsically untrainable, but their optimisation is less robust under the training conditions tested here; greater depth may be advantageous when the relevant displacement scales are genuinely wider. 

%The comparison with recent vision-transformer PIV networks is based on published figures rather than models retrained in our pipeline, and their robustness to tracer loss remains untested. 

%%%%%%%%%%%%%%%%%%%%%%%%%%%%%%%%%%%%%%%%%%%%%%%%%%%%%%%%%%%%%%%%%%%%%
%% METHODS                                                          %
%% NatComm: Methods at the end. Limit 3000 words; no subsection                                      %
%%%%%%%%%%%%%%%%%%%%%%%%%%%%%%%%%%%%%%%%%%%%%%%%%%%%%%%%%%%%%%%%%%%%%
\section*{Methods}
\label{sec:methods}

\subsection*{LIMA architecture}
\label{sec:lima_arch}
% REVISED v2 (Pass 3 / Methods)
$\LIMA$~\cite{manickathan2023lightweight,mucignat2023lightweight,Mucignat2025}
estimates the displacement field, $\mathbf{d}$, from a pair of grey-scale
particle images (\rev{Figure~\ref{fig:LIMA}a}) using an iterative pyramidal architecture
(\rev{Figure~\ref{fig:LIMA}b}). A Siamese encoder, $\mathcal{E}$, shared between
the two input frames, produces feature pyramids that halve the spatial
dimensions at each level. At every level $n$, the features of both frames are 
warped symmetrically by the previous-level displacement estimate, ${\bf d}_{n-1}$, using a second-order
temporal integration~\cite{Wereley2001}. The encoder feature maps of the
warped second frame and of the original first frame are then correlated, and the
resulting correlation volume is passed to the flow decoder $\mathcal{D}$.
The decoder uses dilated convolutions to expand its receptive field
without further downsampling. The displacement estimate at level $n$ is
bilinearly upsampled and used to warp the level-$(n{+}1)$ features, and the
process repeats up the pyramid, in the spirit of the iterative residual
refinement of Hur \textit{et al.}~\cite{hur2019iterative}. Because cross-correlation is
computed on progressively finer feature maps from the
deepest level up to the input resolution, the displacement field reconstructions can be extracted at intermediate levels
to provide a controllable trade-off between accuracy and computational cost. The
present work introduces two new components within this architecture (marked in
grey in Figure~\ref{fig:LIMA}b): a correlation-volume normalisation layer, $\mathcal{N}$, and a two-channel feature projector, $\mathcal{P}$. Both are inserted before the decoder and are described in detail in the Architecture design section.

\subsection*{End-to-end receptive field}
\label{sec:end2end_rf}

The end-to-end spatial support of a single output displacement
is determined by three contributions: the encoder receptive field, the correlation search range, and the
decoder receptive field. We derive these contributions in turn and combine them to obtain the closed-form
expression reported for the end-to-end spatial support in Equation~\eqref{eq:Ltot}.

\emph{Encoder.} The six-level baseline encoder has twelve $3{\times}3$ convolutional layers organised in six stages, each consisting of a strided convolution that halves the spatial dimensions followed by a non-strided convolution of equal kernel size (Table~\ref{tab:feature_extractor_L6}). Each encoder stage contains two convolutions; hence, the output of stage $n$ corresponds to convolution $m=2n$. The number of feature channels increases from 16 at the first stage to 196 at the sixth stage. Two quantities govern the spatial extent over which a single feature element aggregates information: the cumulative stride $ S_m=\prod_{i=1}^{m}s_i, $ which gives the input-to-feature downsampling factor after convolution $m$, and the theoretical receptive field, $\mathrm{RF}_m$, which gives the side length of the input-image region that can influence a single feature element after convolution $m$. At the convolution level, the cumulative stride and receptive field grow recursively as 
\begin{equation} \label{eq:receptive_field_encoder} \begin{aligned} S_m &= S_{m-1}s_m, &\qquad \text{with } S_0 &= 1,\\ \mathrm{RF}_m &= \mathrm{RF}_{m-1} +(k_m-1)d_mS_{m-1}, &\qquad \text{with } \mathrm{RF}_0 &= 1, \end{aligned} 
\end{equation} 
where $k_m$, $d_m$, and $s_m$ denote the kernel size, dilation, and stride of convolution $m$, respectively. \Cref{tab:feature_extractor_L6} reports the cumulative stride, theoretical receptive field and total required padding, $p_{tot}$, after each convolution, calculated using \cref{eq:receptive_field_encoder}. In the subsequent stage-level analysis, $S_n$ and $\mathrm{RF}_n$ denote the corresponding values at the output of the second convolution of stage $n$, \textit{i.e.}, the values evaluated at $m=2n$. The theoretical receptive field is an upper bound; the effective receptive field is typically smaller and depends on the learned network parameters and training distribution~\cite{Luo2017}. The deep pyramid also requires boundary padding corresponding to up to 189 px in input-image space, which can induce reconstruction artefacts, as investigated in~\cite{Mucignat2025}.

\emph{Correlation.} At each pyramid level $n$, the correlation operator
constructs a four-dimensional cost volume of size
$(2 \cdot \mathrm{SR}{+}1) \times (2 \cdot \mathrm{SR}{+}1) \times H_n \times W_n$, where $H_n$ and
$W_n$ are the dimensions of the feature maps and $\mathrm{SR}$ is the fixed search
range. For each spatial location $(x,y)$ and integer offset
$(\Delta_x,\Delta_y) \in [-\mathrm{SR},\mathrm{SR}]^2$, the cost-volume element is the
channel-wise inner product
\begin{equation}
C_n(x,y,\Delta_x,\Delta_y)
= \sum_{c=1}^{C_n} \tilde{F}^{(1)}(x,y,c)\,
                   \tilde{F}^{(2)}(x{+}\Delta_x,\,y{+}\Delta_y,\,c),
\label{eq:corr}
\end{equation}
where $\tilde{F}^{(1)}$ and $\tilde{F}^{(2)}$ denote the warped feature maps at level $n$, and
$C_n$ is the channel dimension. A feature-space shift of $\Delta$
corresponds to an input-space translation of $\Delta\,S_n$ pixels, such that
the correlation operator acts as an effective $(2 \cdot \mathrm{SR}{+}1)$-wide aggregation kernel in
feature space and the input-space support of a single
correlation output is
\begin{equation*}
L_{\mathrm{corr}} \;=\; \mathrm{RF}_n + 2 \cdot \mathrm{SR} \cdot S_n.
\end{equation*}
For the six-level encoder with $\mathrm{SR} = 2$, the maximum input-space
displacement represented by the correlation search window reaches $\mathrm{SR} \cdot S_6
= 128$~px at the deepest level (Table~\ref{tab:end2end_rf}). For the targeted PIV displacement regime, which extends to approximately $20$\,px in the present experiments, three pyramid levels provide sufficient search capacity through the combination of a cumulative stride $S_3=8$, a correlation range $\mathrm{SR}=2$, and coarse-to-fine warping with residual refinement.
Deeper levels only enlarge the spatial context in which the displacement is interpreted but do not provide additional benefits for the reconstruction.

\emph{Decoder and end-to-end support.} The decoder is a stack of dilated $3{\times}3$
convolutions with feature-space receptive field
$\mathrm{RF}_{\mathrm{dec}}$ (see Table~\ref{tab:context_network_L6} for the
six-level baseline). Because the decoder operates at unit stride in feature space, it
extends the input-space support of a single output element by
$(\mathrm{RF}_{\mathrm{dec}} - 1)\,S_n$. Combining the encoder, correlation,
and decoder yields the end-to-end receptive fields in Equation~\eqref{eq:Ltot},
\begin{equation}
L_{\mathrm{total}}(n)
\;=\;
\underbrace{\mathrm{RF}_n + 2 \cdot \mathrm{SR} \cdot S_n}_{\text{correlation span}}
\;+\;
\underbrace{(\mathrm{RF}_{\mathrm{dec}} - 1)\,S_n}_{\text{decoder enlargement}}
\;=\;
\mathrm{RF}_n + \bigl(2 \cdot \mathrm{SR} + \mathrm{RF}_{\mathrm{dec}} - 1\bigr)\,S_n.
\label{eq:Ltot_methods}
\end{equation}
For the six-level encoder with $\mathrm{SR} = 2$ and a 67-pixels decoder, this
reduces to $L_{\mathrm{total}} = \mathrm{RF}_n + 70 \cdot S_n$ in input
pixels. The corresponding values per pyramid level are reported in
Table~\ref{tab:end2end_rf}, with $L_{\mathrm{total}} = 4{,}859$~px
at level~6. This quantitative result motivates the
architectural redesign in the next subsection: at the deepest layer,
the end-to-end support exceeds the physically meaningful displacement range by two orders of magnitude. This additional spatial reach is not required to represent the targeted displacement and incurs additional stride, padding, memory, and computational cost.

\begin{table}[h!]
\centering
\caption{\textbf{Six-level feature extractor}. Kernel configuration, cumulative stride, theoretical receptive field, feature-map size, and cumulative input-space padding, assuming an input image of $256\times256$~px.}
\label{tab:feature_extractor_L6}
\begin{tabular}{lcccccccccccc}
\hline
Layer & $m$ & $c_{in}$ & $c_{out}$ & $k$ & $s$ & $S_m$ & $d$ & $p$ & $\mathrm{RF}$ & Feat. size & $p_{tot}$ \\
\hline
1.0 & 1  &  1   &  16  & $3\times3$ & 2 &  2 & 1 & 1 &   3 & $128\times128$ & -- \\
1.1 & 2  & 16   &  16  & $3\times3$ & 1 &  2 & 1 & 1 &   7 & $128\times128$ &  3 \\
2.0 & 3  & 16   &  32  & $3\times3$ & 2 &  4 & 1 & 1 &  11 & $ 64\times 64$ & -- \\
2.1 & 4  & 32   &  32  & $3\times3$ & 1 &  4 & 1 & 1 &  19 & $ 64\times 64$ &  9 \\
3.0 & 5  & 32   &  64  & $3\times3$ & 2 &  8 & 1 & 1 &  27 & $ 32\times 32$ & -- \\
3.1 & 6  & 64   &  64  & $3\times3$ & 1 &  8 & 1 & 1 &  43 & $ 32\times 32$ & 21 \\
4.0 & 7  & 64   &  96  & $3\times3$ & 2 & 16 & 1 & 1 &  59 & $ 16\times 16$ & -- \\
4.1 & 8  & 96   &  96  & $3\times3$ & 1 & 16 & 1 & 1 &  91 & $ 16\times 16$ & 45 \\
5.0 & 9  & 96   & 128  & $3\times3$ & 2 & 32 & 1 & 1 & 123 & $  8\times  8$ & -- \\
5.1 & 10 & 128  & 128  & $3\times3$ & 1 & 32 & 1 & 1 & 187 & $  8\times  8$ & 93 \\
6.0 & 11 & 128  & 196  & $3\times3$ & 2 & 64 & 1 & 1 & 259 & $  4\times  4$ & -- \\
6.1 & 12 & 196  & 196  & $3\times3$ & 1 & 64 & 1 & 1 & 379 & $  4\times  4$ & 189 \\
\hline
\end{tabular}
\end{table}

\begin{table}[h!]
\centering
\caption{\textbf{End-to-end receptive field for the ``$\mathrm{SR}=2$ correlation $\to$
67-RF decoder'' six-level baseline}. Per-level cumulative stride $S_n$,
encoder receptive field $\mathrm{RF}_n$, correlation span
$L_{\mathrm{corr}} = \mathrm{RF}_n + 4\,S_n$, decoder enlargement
$L_{\mathrm{dec}} = 66\,S_n$, and total
$L_{\mathrm{total}} = \mathrm{RF}_n + 70\,S_n$. All quantities in
input-space pixels.}
\label{tab:end2end_rf}
\begin{tabular}{lccccc}
\hline
Layer & $\mathrm{RF}_n$ & $S_n$ & $L_{\mathrm{corr}}=\mathrm{RF}_n+4S_n$ &
$L_{\mathrm{dec}}=66S_n$ & $L_{\mathrm{total}}=\mathrm{RF}_n+70S_n$ \\
\hline
1 &   7 &  2 &  15 &  132 &  147 \\
2 &  19 &  4 &  35 &  264 &  299 \\
3 &  43 &  8 &  75 &  528 &  603 \\
4 &  91 & 16 & 155 & 1056 & 1211 \\
5 & 187 & 32 & 315 & 2112 & 2427 \\
6 & 379 & 64 & 635 & 4224 & 4859 \\
\hline
\end{tabular}
\end{table}

\begin{table}[h]
    \centering
    \caption{\textbf{The baseline six-level decoder and
    receptive-field growth}. The total required padding is given in
    feature-space coordinates.}
    \label{tab:context_network_L6}
    \begin{tabular}{lccccccc}
        \hline
        Layer & $c_{in}$ & $c_{out}$ & $k$ & $s$ & $d$ & $\mathrm{RF}$ & $p_{\text{dec}}$ \\
        \hline
        1 & 27  & 128 & $3\times3$ & 1 & 1  & 3  & -- \\
        2 & 128 & 128 & $3\times3$ & 1 & 2  & 7  & -- \\
        3 & 128 & 128 & $3\times3$ & 1 & 4  & 15 & -- \\
        4 & 128 & 96  & $3\times3$ & 1 & 8  & 31 & -- \\
        5 & 96  & 64  & $3\times3$ & 1 & 16 & 63 & -- \\
        6 & 64  & 32  & $3\times3$ & 1 & 1  & 65 & -- \\
        7 & 32  & 2   & $3\times3$ & 1 & 1  & \textbf{67} & \textbf{33} \\
        \hline
    \end{tabular}
\end{table}

\subsection*{Architecture  design}
\label{sec:lima3_design}
Three modifications to the baseline architecture are introduced as direct consequences of the
receptive-field estimate in Equation~\eqref{eq:Ltot_methods}. First,
the encoder depth is reduced from six to three levels, eliminating the
three deepest pyramid stages whose spatial supports substantially exceed the
relevant displacement range. The resulting three-level encoder retains the
strided-plus-non-strided convolution pattern of the original, with a
$5{\times}5$ kernel in the second sub-stage of the deepest level
(Table~\ref{tab:network_overview}, panel~\subref{tab:feature_extractor})
to extend the deepest-layer receptive field without further
downsampling. Second, the decoder receptive field is reduced from
67 to 15 by dropping the dilation pattern of the original context
network in favour of uniform $3{\times}3$ convolution kernels with dilation~$1$
(panel~\subref{tab:context_network}). Third, two lightweight modules
are inserted between the correlation operator and the decoder: a
correlation-volume normalisation layer that rescales correlation
responses to a fixed dynamic range, and a two-channel feature projector that compresses the concatenated encoded features  from $2C_n$ channels to 2, which are provided to the decoder alongside the cost volume and the displacement estimate from the previous pyramid level. 

The architectural modifications reduce the end-to-end receptive field of $\LIMA{^{3}}$
(panel~\subref{tab:LIMAV3}) to $L_{\mathrm{total}} \leq 203$~px at
the deepest level, the maximum required input padding from $189$~px to
$29$~px, and the cumulative stride at the deepest layer to
$S_3 = 8$, thereby allowing the loss to be evaluated on denser spatial grids than
in the six-level baseline.

The number of parameters is $137{,}000$ for the three-level encoder, $18$ for
the correlation-normalisation layer, $904$ for the feature projector,
and $451{,}000$ for the decoder (the warping operator has no learnable
weights), for a total of about $589{,}000$, a reduction from approximately 
$1.6$~million for $\LIMA{^{6,0}}$~\cite{Mucignat2025}, which is also substantially smaller
than $\RAFT$~\cite{teed2020raft} ($5.3$~M),
LiteFlowNet~\cite{hui2018liteflownet} ($5.37$~M),
FlowNetS~\cite{dosovitskiy2015flownet} ($38$~M), and
FlowNet2~\cite{ilg2017flownet} ($162.5$~M).

We train three variants, denoted
$\LIMA{^{3,\ell}}$ with $\ell \in \{0,1,2\}$, that differ only in the pyramid level at
which reconstruction is terminated. The variants share weights up to their respective termination
levels. Hence, $\ell = 0$ produces full-resolution reconstructions 
at the input resolution, whereas $\ell = 1$ and $\ell = 2$ terminate at the first
and second encoding levels, yielding displacement fields at half and quarter resolution, respectively.

\begin{table}[h!]
\centering
\caption{\textbf{$\LIMA{^{3}}$ architecture}. Parameters and end-to-end receptive field analysis.}
\label{tab:network_overview}

\begin{subtable}{\textwidth}
\centering
\caption{Feature extractor with 3 layers, assuming an input image of
$256\times256$~px.}
\label{tab:feature_extractor}

\begin{tabular}{lccccccccccc}
\toprule
Layer & $m$ & $c_{in}$ & $c_{out}$ & $k$ & $s$ & $S_m$ & $d$ & $p$ & $\mathrm{RF}$ & Feat.\ size & $p_{tot}$ \\
\midrule
1.0& 1 &  1 & 16 & $3\times3$ & 2 & 2 & 1 & 1 &  3 & $128\times128$ & --\\
1.1& 2 & 16 & 16 & $3\times3$ & 1 & 2 & 1 & 1 &  7 & $128\times128$ &  3\\
\addlinespace
2.0& 3 & 16 & 32 & $3\times3$ & 2 & 4 & 1 & 1 & 11 & $ 64\times 64$ & --\\
2.1& 4 & 32 & 32 & $3\times3$ & 1 & 4 & 1 & 1 & 19 & $ 64\times 64$ &  9\\
\addlinespace
3.0& 5 & 32 & 64 & $3\times3$ & 2 & 8 & 1 & 1 & 27 & $ 32\times 32$ & --\\
3.1& 6 & 64 & 64 & $5\times5$ & 1 & 8 & 1 & 2 & 59 & $ 32\times 32$ & 29\\
\bottomrule
\end{tabular}
\end{subtable}

\vspace{0.8em}

\begin{subtable}{\textwidth}
\centering
\caption{Context network for $\LIMA{^{3}}$ and receptive-field growth.
Total required padding is given in feature-space coordinates.}
\label{tab:context_network}
\begin{tabular}{lccccccc}
\toprule
Layer & $c_{in}$ & $c_{out}$ & $k$ & $s$ & $d$ & $\mathrm{RF}$ & $p_{\text{feature}}$\\
\midrule
1 & 29  & 128 & $3\times3$ & 1 & 1 &  3 & 1\\
2 & 128 & 128 & $3\times3$ & 1 & 1 &  5 & 2\\
3 & 128 & 128 & $3\times3$ & 1 & 1 &  7 & 3\\
4 & 128 &  96 & $3\times3$ & 1 & 1 &  9 & 4\\
5 &  96 &  64 & $3\times3$ & 1 & 1 & 11 & 5\\
6 &  64 &  32 & $3\times3$ & 1 & 1 & 13 & 6\\
7 &  32 &   2 & $3\times3$ & 1 & 1 & \textbf{15} & \textbf{7}\\
\bottomrule
\end{tabular}
\end{subtable}

\vspace{0.8em}

\begin{subtable}{\textwidth}
\centering
\caption{End-to-end receptive field for $\LIMA{^{3}}$
($\mathrm{SR}=2$ correlation followed by a 15-RF decoder). All values in input-space
pixels.}
\label{tab:LIMAV3}
\begin{tabular}{lccccc}
\toprule
Layer & $\mathrm{RF}_n$ & $S_n$ & $L_{\mathrm{corr}}=\mathrm{RF}_n+4S_n$ & $L_{\mathrm{dec}}=14S_n$ & $\mathbf{L_{\mathrm{total}}=\mathrm{RF}_n+18S_n}$ \\
\midrule
1 &  7 & 2 & $ 7 + 8 = 15$  & $ 28$ & $\mathbf{ 43}$\\
2 & 19 & 4 & $19 +16 = 35$  & $ 56$ & $\mathbf{ 91}$\\
3 & 59 & 8 & $59 +32 = 91$  & $112$ & $\mathbf{203}$\\
\bottomrule
\end{tabular}
\end{subtable}

\end{table}

\subsection*{Synthetic training data}
\label{sec:training_data}
Training data are obtained with the kinematic
generator introduced in~\cite{manickathan2022kinematic} and further
extended in the present work, with image formation based on our open-source synthetic
generator~\cite{Zdyba2025}. The pipeline builds each image pair by seeding 
particles at random initial positions, displacing them with a randomly 
parameterised flow field, and rendering the two frames under randomly drawn 
imaging and experimental-setting parameters. This avoids the need for direct 
numerical simulations and enables inexpensive generation of effectively 
unlimited training data. The flow, imaging, and experimental-setting 
parameters are drawn independently for each image pair from the distributions 
in Table~\ref{tab:params}. The horizontal displacement magnitude is sampled
log-uniformly from $[1.5, 20]$~px, with $17\%$ of the samples explicitly
allocated to the sub-pixels regime $[0,1.5]$. \rev{The smoothing scale, which
controls the spatial-frequency content of the flow, is sampled 
from $[3, 155]$~px.} Seeding density is sampled
uniformly from $[0.02, 0.18]$ particles-per-pixel, particle diameters and
shapes are sampled for each image pair, and the laser-sheet illumination profile
is modelled as a Gaussian with randomly sampled parameters.

Particle advection through the synthetic flow field is performed using a fourth-order
Runge-Kutta integrator with the standard $(1/6, 1/3, 1/3, 1/6)$
weights and cubic interpolation for off-grid velocity sampling. The
implementation is validated against six numerical correctness tests, all passing at machine-precision tolerances: exact reproduction of constant-flow trajectories, linear time-step scaling, exact reproduction of pure-shear trajectories, bounded fifth-order error on solid-body rotation, fourth-order accuracy under time-step halving, and out-of-domain particle flagging.

\paragraph{Out-of-plane particle loss model.}
Out-of-plane particle loss is a major inter-frame degradation mechanism
that distinguishes PIV from natural-image optical flow;
it is modelled from first principles rather than as an \textit{ad hoc} gating mechanism. For a
three-dimensional, incompressible flow, the continuity equation gives
$\nabla \cdot \boldsymbol{u}_{3D} = 0$, which
implies $\nabla \cdot \boldsymbol{u}_{2D} = -\partial w / \partial z$ when the flow is restricted to the laser sheet.
The in-plane divergence provides a per-pixel proxy for the fraction of
particles leaving the laser sheet between frames. The
divergence-coupled per-pixel loss term is
\begin{equation}
\Pi_{\mathrm{div},(i,j)}
= \min\!\bigl( \, |\nabla \cdot \boldsymbol{u}_{2D}|_{(i,j)} \, \Delta t, \; 1\bigr),
\label{eq:oop_div}
\end{equation}
\rev{to which we add a random component $\Pi_{\mathrm{rnd}}$ drawn from a
triangular distribution on $[0, \Pi_{\mathrm{cap}}]$ with its peak at zero,
accounting for unresolved 3D structures not represented in the
two-dimensional flow. The total loss is clipped at $\Pi_{\mathrm{cap}} =
0.50$, keeping most pairs within the PIV guidelines of Keane and
Adrian~\cite{KeaneAdrian1992}.}

\paragraph{Fully-online generation.}
Each training example is synthesised on the GPU at every step. 
This process replaces storage of a pre-rendered set with on-the-fly generation. All per-image-pair parameters (particle density, diameter, eccentricity, brightness,
laser-sheet profile, the random and divergence-coupled out-of-plane
components, the buffer crop, and the augmentation chain) are sampled
independently for each example, ensuring that the network never sees the same image pair twice and yielding an effectively unbounded, rich training set. 
\begin{table}[h]
\footnotesize
\centering
\caption{\textbf{Parameter distributions for the synthetic training
generator.}}
\label{tab:params}
\begin{tabular}{lcc}
\hline
Quantity & Distribution & Role \\
\hline
Displacement $\|\mathbf{u}\|$ & log-uniform $[1.5, 20]$~px (+17\% sub-px) & dynamic range \\
Smoothing scale $\sigma_f$ & $[3, 155]$~px  & spatial-frequency content \\
Seeding density $\rho$ & uniform $[0.02, 0.18]$~part./px & image texture \\
Particle diameter, shape & per-image draw & image formation \\
Laser-sheet Gaussian shape & per-image draw & illumination profile \\
OOP random component & \rev{triangular $[0,\Pi_{\mathrm{cap}}{=}0.50]$, peak at $0$} & unresolved 3D structure \\
OOP physical component & $|\nabla \cdot \boldsymbol{u}_{2D}| \Delta t$ (Eq.~\ref{eq:oop_div}) & continuity-derived loss \\
\hline
\end{tabular}
\end{table}

\subsection*{Training procedure}
\label{sec:training}
All $\LIMA{^{3}}$ variants are trained end-to-end from scratch using the AdamW
optimiser with default parameters $(\beta_1, \beta_2)$ and an initial learning
rate of $4{\times}10^{-4}$, reduced on a validation-loss plateau
(\texttt{ReduceLROnPlateau}, patience 7, minimum learning rate
$10^{-8}$). Weights are initialised by sampling from a
Kaiming distribution. For the
factorial ablation study, each of the
sixteen variants is trained independently under identical
hyper-parameters to isolate architectural effects\rev{, with a base learning rate of $2{\times}10^{-4}$ reduced on plateau until a common floor of $3.1{\times}10^{-6}$ is reached (151--240~epochs). At each pyramid level the displacement update of the decoder is bounded to the search range, $|\Delta\mathbf{d}|\le\mathrm{SR}$. The three-level configurations use a hard bound, whereas the six-level configurations use the softsign bound $\Delta\mathbf{d}=\mathbf{x}/(1+|\mathbf{x}|/\mathrm{SR})$, whose gradient decays as $(1+|\mathbf{x}|/\mathrm{SR})^{-2}$ rather than vanishing (hard bound) or decaying exponentially ($\mathrm{SR}\tanh(\mathbf{x}/\mathrm{SR})$)}. The
four reported production checkpoints are trained for
up to 200~epochs, with the checkpoint selected on the basis of the endpoint error on a held-out validation set of $1{,}024$ samples (disjoint from both training and test sets). All selected checkpoints plateau well before
this ceiling. Training and inference are performed on the same hardware, a single
NVIDIA RTX~5090, and training proceeds at approximately $17$~minutes per epoch
($\sim$19~ms per image pair).

\subsection*{Domain-randomisation augmentation}
\label{sec:augmentation}
Generalisation to experimental recordings is supported by an
extensive online augmentation pipeline applied at each training step,
following the domain-randomisation framework of~\cite{Tobin2017}. The
pipeline comprises five groups of stochastic transformations
(geometric flips/transpose, intensity-dynamics rescaling, ambient and
laser-sheet background, intensity-dependent Poisson shot noise, and
mask injection for laser reflections, laser stripes, and opaque
occluders).

The full specification of each primitive (sampling ranges, \rev{mask-injection}
primitives, laser-stripe halo/speckle) is
provided in the Supplementary Methods.

\subsection*{Loss function}
\label{sec:loss_function}
% REVISED v2 (Pass 3 / Methods)
The training loss combines a displacement-fidelity term with \rev{a
gradient-consistency regulariser}, applied at each pyramid
level. The displacement-fidelity term uses the Charbonnier penalty
\begin{equation}
L_{\mathrm{disp}} = \frac{1}{N} \sum
\Bigl(\sqrt{(u - u_{\mathrm{GT}})^2 + \epsilon^2}
    + \sqrt{(v - v_{\mathrm{GT}})^2 + \epsilon^2}\Bigr),
\label{eq:Ldisp}
\end{equation}
with $\epsilon = 10^{-3}$, where $(u, v)$ and $(u_{\mathrm{GT}},
v_{\mathrm{GT}})$ denote the \rev{estimated} and ground-truth displacement fields,
and $N$ is the number of valid pixels in the batch. \rev{An additional
regularisation term, introduced in earlier $\LIMA$
versions~\cite{manickathan2023lightweight,mucignat2023lightweight}, is
defined from the Jacobian $\boldsymbol{J}$ of the displacement field and
penalises the pixel-wise Frobenius norm of the Jacobian difference,
$L_{\mathrm{Fro}} = \frac{1}{N} \sum \sqrt{\|\boldsymbol{J} - \boldsymbol{J}_{\mathrm{GT}}\|_{F}^{2} + \epsilon^{2}}$,
matching all four velocity-gradient components
$(\partial u/\partial x,\ \partial u/\partial y,\ \partial v/\partial x,\ \partial v/\partial y)$
to the ground truth.} The total loss at pyramid level
$i$ is a weighted sum
\rev{$L_i = \lambda_1 L_{\mathrm{disp}} + \lambda_2 L_{\mathrm{Fro}}$,
with $\lambda_1 = 1$ and $\lambda_2 = 0.05$.} The overall
loss is the weighted sum over the pyramid
levels, $L = \sum_i w_i L_i$, with weights emphasising the final reconstruction level.

\subsection*{Evaluation protocol and error metrics}
\label{sec:eval_protocol}
% REVISED v2 (Pass 3 / Methods)
 \LIMA{}, \RAFT{} and \WIDIM{} shared the same image pre-processing pipeline, comprising a sliding minimum subtraction; for the learned methods an additional per-pair normalisation was required upfront.
On cases where ground-truth displacement fields are available, accuracy is quantified by the endpoint error
$\epsilon = \sqrt{(u - u_{\mathrm{GT}})^2 + (v - v_{\mathrm{GT}})^2}$. As \LIMA{$^{3,1}$}, \LIMA{$^{3,2}$}, and \WIDIM{} produce vector fields at reduced resolution we down-sample the ground truth at the vector positions using a bi-cubic interpolator.
The displacement curl $\zeta = \partial v/\partial x - \partial u/\partial y$ is proportional to the physical out-of-plane vorticity, the two differing only by the constant inter-frame time step; its reconstruction accuracy is quantified analogously by the per-pixels displacement-curl error $\epsilon_\zeta = \lvert \zeta  - \zeta_{\mathrm{GT}} \rvert$, where $\zeta_{\mathrm{GT}} = \partial v_{\mathrm{GT}}/\partial x - \partial u_{\mathrm{GT}}/\partial y$ is the curl of the ground-truth displacement field. \rev{$\zeta$ is evaluated by central differences on the vector grid of each method, and $\zeta_{\mathrm{GT}}$ is evaluated exactly (spectrally, as the DNS field is periodic) at the same vector positions.}

Spatial resolution is quantified on the sinusoidal-shear-flow benchmark, in which the imposed horizontal displacement $u = A_0\sin(2\pi y/\lambda)$ is swept in wavelength.
At each $\lambda$ the reconstructed field is projected onto the imposed sinusoid to give the amplitude response $R_A(\lambda) = A/A_0$. Expressed as a function of the spatial frequency $f = 1/\lambda$, the sweep constitutes the modulation transfer
function of an equivalent linear filter, and the equivalent filter length $L_{sr}$ is its equivalent width~\cite{BernhardWieneke2017},
\begin{equation}
  L_{sr} \;=\; \frac{R_A^{2}(f \! \to \! 0)}
                    {2\displaystyle\int_{0}^{f_{\max}} R_A^{2}(f)\,\mathrm{d}f},
  \label{eq:lsr}
\end{equation}
where $R_A(f \! \to \! 0)$ denotes the response at the longest resolved wavelength.
For an ideal moving-average filter Eq.~\eqref{eq:lsr} returns exactly the kernel width, so that $L_{sr}$ is directly interpretable as a length in pixels and smaller
values indicate finer resolution. Because the entire response curve enters the denominator, an amplitude response
exceeding unity at intermediate wavelengths lowers $L_{sr}$ even where the reconstruction has degraded; such points are flagged and excluded from the analysis. Note that $L_{sr}$ is evaluated numerically with a discrete cosine transform over  $f \in [0,1]$ at the wavelength increment $\Delta\lambda = 1.37$~px.

For experimental cases without ground truth, reconstruction
quality is assessed by two complementary approaches. 
The first relies on the disparity
error $\epsilon_p$~\cite{Sciacchitano2013}: the reconstructed 
displacement field is applied symmetrically to advect first-frame 
particles half-way forward and second-frame particles half-way 
backward, and the residual offset between the particle images is computed.
The second comprises three additional matching-free indicators computed on the
symmetrically warped image pair: the correlation peak ratio (PR),
defined as the ratio of the primary to the secondary correlation peak;
the primary correlation coefficient $c$, normalised to $[0,1]$; and the
residual displacement magnitude $|d_{\mathrm{res}}|$, which is the residual
misalignment of the warped pair measured by correlation rather than 
particle matching. A high-quality reconstruction yields
large PR and $c$ together with a tightly peaked, near-zero
$|d_{\mathrm{res}}|$ distribution.

\subsection*{Audit of public training data}
\label{sec:set_audit}
The two physical-fidelity gaps examined in the Discussion were
quantified directly from the released image pairs and ground-truth
displacement fields of the Cai 2019 dataset~\cite{Cai2019} and the
\RAFT{} Problem Class 2 deposit~\cite{lagemann2021deep}. As neither
original generator is publicly available, the audit is restricted to
the distributed data. The two datasets advect particles with a
symmetric advection scheme \cite{Cai2019} or fourth-order
Runge--Kutta integrator \cite{lagemann2022generalization}; the two agree to within $\sim\!0.001$~px
for a  displacement of 20 px, well below PIV measurement noise;
the advection scheme is therefore not a meaningful source of fidelity
loss.

The picture is different in the case of particle loss: tracer particles in the first frame may have no match in the second because out-of-plane motion carries them out of the laser sheet. 
We tested both datasets using only released
image pairs and ground-truth displacement, with estimators based on
local brightness conservation, divergence coupling, and direct
particle-by-particle pairing, each calibrated against a synthetic
no-loss reference at matched particle, contrast and noise statistics
and validated on injected-loss controls. Under these diagnostics, we detected no evidence of particle loss in
either dataset. The  PC2 data carry a large apparent
decorrelation, but it lies below the dataset's own no-loss reference
and is consistent with heavy imaging noise and out-of-plane intensity
modulation rather than lost particles: a particle-matching
test finds the displaced first-frame particles still land on real
second-frame particles, with sub-pixels residuals.

\subsection*{\RAFT{} processing pipeline}
\label{sec:RAFT_eval}
For comparison against state-of-the-art learned baselines we \rev{evaluated} \RAFT{$_{32}$}  and \RAFT{$_{256}$}, using the pretrained models obtained from the associated Code Ocean capsule \cite{codeocean_RAFT} and run in ONNX form. \RAFT{$_{32}$} employs a restricted $32\times32$ pixels interrogation window and performs feature extraction without spatial downsampling, whereas \RAFT{$_{256}$} operates on the full $256\times256$ pixels frame, downsampling to one-eighth resolution during feature extraction before convex upsampling back to a dense field; both produce per-pixels displacement estimates through iterative all-pairs correlation updates (12 and 16 GRU iterations, respectively). For both networks, input image pairs were normalized to the $[0, 1]$ range using the same per-pair normalisation applied to \LIMA{}, and all networks were applied in a single whole-frame forward pass. No post-processing was applied to the reconstructed displacement fields.

\subsection*{\WIDIM{}{} processing pipeline}
\label{sec:WIDIM_eval}
We used a commercial state-of-the-art implementation of \WIDIM{} (\emph{Davis}, LaVision) with an initial interrogation window of $64\times64$ pixels \rev{and} six passes at the final interrogation window of $8,16,24$ or $32$ pixels depending on the specific case. For the final interrogation windows we used a gaussian weighting to improve resolution of the vector fields \cite{Astarita2007}. Grid overlap was set to $75\%$ for all sets, and validation was performed based on a Normalized Median Filter (NMF) on a $5\times5$ kernel followed by one pass of anisotropic denoising \cite{Wieneke2017} on the final fields. 
%%%%%%%%%%%%%%%%%%%%%%%%%%%%%%%%%%%%%%%%%%%%%%%%%%%%%%%%%%%%%%%%%%%%%

\section*{Funding}
We acknowledge funding from the Swiss National Science Foundation under the project AuRos (Grant 10005849) and access to Swiss National Supercomputing Centre (CSCS), Switzerland, under Empa’s share with the project ID em13.

\section*{Acknowledgements}
We thank Stephan Kunz for assistance with the code, and Hossein \rev{Gorji} for fruitful discussions.
C.M., K.Z., and I.L. disclose support from internal Empa research funds
for the work described in this study.

\section*{Author contributions}
C.M.: conceptualisation, methodology, software, validation, investigation,
visualisation, writing, original draft.
K.Z.: conceptualisation, software, writing, review and editing.
I.L.: conceptualisation, resources, writing, review and editing.

\section*{Competing interests}
The authors declare no competing interests.

\section*{Data availability}
The Problem Class 1 and 2 datasets are available online at \cite{cai_piv_dataset_github} and \cite{codeocean_RAFT}.
The DNS and analytical flow datasets used for benchmarking are publicly
available at ~\cite{CarlierDNS,Carlier}. PIV challenge datasets can be found at \cite{PIVChallenge}. 
No training dataset is deposited because $\LIMA{^{3}}$
is trained entirely from a fully online, synthetic
data generator available at \cite{Zdyba2025}.
%Experimental datasets are available at the \LIMA{} repository \cite{LIMA3} or at \cite{codeocean_RAFT}.

%\section*{Code availability}
%The trained \rev{\LIMA{} models are available} at \cite{LIMA3}. Provisional link to access the Zenodo repository during the review:\url{ https://zenodo.org/records/22043147?preview=1&token=eyJhbGciOiJIUzUxMiJ9.eyJpZCI6ImNkZGIyMTI0LTE4ZGUtNDIxNC04NjQyLTdiNDE2MzE2M2U1YiIsImRhdGEiOnt9LCJyYW5kb20iOiJlMjRmM2IyOGY0ODk1M2ZmNjE3YzE2N2Y1NDE5MThmNiJ9.TGZ8aTfKBqOpekaNH5rQ8IL6yztQarX7Lb_sc5Ibm82qmpxFhYfJoiODD8gDvO5tlGpDqvn4_NChBvsbMTIKfw}.

\bibliographystyle{naturemag}
\bibliography{bibliography}

\clearpage

\setcounter{figure}{0}
\setcounter{table}{0}
\renewcommand{\figurename}{Supplementary Figure}
\renewcommand{\tablename}{Supplementary Table}

\begin{center}
{\Large\textbf{Supplementary Information}}
\end{center}

\section*{Domain-randomisation augmentation (full specification)}
\label{sec:augmentation_supp}
The online augmentation pipeline organises stochastic transformations
into five groups.

\emph{Geometric transformations} cover horizontal flip, vertical flip,
and transpose, with a $5\%$ probability substitution of a
zero-displacement input pair that explicitly teaches
zero-reconstruction.

\emph{Intensity-dynamics transformations} model variation in camera
dynamic range and exposure: per-sample integer count
rescaling uniform in $[100, 10{,}000]$, per-frame brightness
scaling uniform in $[0.5, 1.5]$ (emulating laser pulse-to-pulse energy jitter), and a sinusoidal spatial-gain
modulation.

\emph{Background transformations} emulate ambient illumination and
non-uniform laser-sheet brightness: a uniform offset in $[0, 300]$
counts and a Gaussian-blob spatial background applied to $50\%$ of
samples.

\emph{Photon shot noise} is modelled as an intensity-dependent
Poisson process: photon counts are drawn at a scale factor sampled
uniformly in $[0.05, 10]$, with low scale producing high relative shot
noise
(approximately $1/\sqrt{N}$) and high scale producing nearly noise-free
images. Quantisation to integer counts follows.

\emph{Mask injection} models laser-reflection halos, laser stripes, and
opaque occluders through a unified shape-injection transform applied to
$25\%$ of samples. One shape is drawn per affected sample from a
uniform menu over five primitives (circle, ellipse, parallelogram,
laser reflection, and laser stripe), each occupying \rev{$20$--$30\%$} of
the image area with fill values drawn uniformly in
$[0, 40{,}000]$~counts (i.e., from fully transparent to
sensor-saturating intensity). Solid occluders are tagged with a
Gaussian edge-glow halo at the boundary with $50\%$ probability
($\sigma = 2.5$~px, decay length $6$~px). Laser reflections are modeled with a halo blur ($\sigma = 3$~px), a high-frequency speckle
modulation (strength $0.6$, $\sigma = 0.7$~px), and a pulse-to-pulse
intensity jitter (standard deviation $0.07$).

\section*{Additional benchmark on Problem Class 1 dataset}

\begin{table}[htbp]
\centering
\footnotesize
\caption{\textbf{Superseded root-mean-square error baselines on Problem Class 1.} Table reproduces values from \LIMA{}, our \RAFT{} and \WIDIM{} evaluations, and  the data from Table I of Ref.~\cite{Cai2019b}, the error metric is defined as the root-mean-square error over all image pixels hence the data are  not comparable with Table~\ref{tab:cai_class1}. Values in px, frame$^{-1}$}
\label{tab:cai_rmse_si}
\begin{tabular}{l ccccc}
\toprule
Method & Backstep & JHTDB-ch & DNS-turb & Cylinder & SQG \\
\midrule
\multicolumn{6}{l}{\itshape Published root-mean-square errors, reproduced from~\cite{Cai2019b}} \\
$\WIDIM$            & 0.034 & 0.084 & 0.304 & 0.083 & 0.457 \\
Horn--Schunck       & 0.045 & 0.069 & 0.135 & 0.070 & 0.156 \\
OF+STE              & 0.047 & 0.078 & 0.110 & 0.079 & 0.110 \\
PIV-DCNN            & 0.049 & 0.117 & 0.334 & 0.100 & 0.479 \\
PIV-FlowNetS        & 0.139 & 0.237 & 0.525 & 0.194 & 0.525 \\
PIV-FlowNetS-en     & 0.072 & 0.155 & 0.282 & 0.115 & 0.294 \\
PIV-LiteFlowNet     & 0.056 & 0.104 & 0.196 & 0.083 & 0.202 \\
PIV-LiteFlowNet-en  & 0.033 & 0.075 & 0.122 & 0.049 & 0.126 \\
\midrule
\multicolumn{6}{l}{\itshape This work, same statistic and same test split} \\
% >>> CHANGED 2026-09-15: same v5 re-score as Table~\ref{tab:cai_class1}.
% >>> CHANGED 2026-09-29: v5b re-score (rmse_sharp_buf32); PC1-control row corrected to the
% >>>          augmented checkpoint (was stale 0.015/0.027/0.052/0.019/0.044).
% >>> CHANGED 2026-10-21: reader half-pixel fix; cai_class1_aee.csv rmse_sharp_buf32.
% >>> CHANGED 2026-10-29: render-1.0 production LIMA rows (cai_class1_aee.csv, rmse_sharp_buf32).
$\LIMA{^{6,0}}$ & \rev{0.034} & \rev{0.113} & \rev{0.133} & \rev{0.065} & \rev{0.142} \\
$\LIMA{^{3,0}}$ & \rev{0.027} & \rev{0.073} & \rev{0.105} & \rev{0.042} & \rev{0.103} \\
$\LIMA{^{3,1}}$ & \rev{0.045} & \rev{0.118} & \rev{0.157} & \rev{0.072} & \rev{0.161} \\
$\LIMA{^{3,2}}$ & \rev{0.069} & \rev{0.204} & \rev{0.280} & \rev{0.137} & \rev{0.290} \\
$\LIMA{^{3,0}_{\mathrm{PC1}}}$ & \rev{0.017} & \rev{0.032} & \rev{0.060} & \rev{0.023} & \rev{0.051} \\
$\WIDIM$ $16\times16$ & 0.019 & 0.064 & 0.247 & 0.048 & 0.254 \\
$\WIDIM$ $8\times8$   & 0.107 & 0.086 & 0.195 & 0.091 & 0.132 \\
$\RAFT{_{256}}$ PC1   & 0.013 & 0.113 & 0.083 & 0.026 & 0.112 \\
$\RAFT{_{32}}$ PC1    & 0.006 & 0.015 & 0.036 & \rev{0.027} & 0.023 \\
\bottomrule
\end{tabular}
\end{table}

The values in the upper block of Supplementary Table~\ref{tab:cai_rmse_si} are root-mean-square errors, as defined in Ref.~\cite{Cai2019b}, and are reproduced identically in Refs.~\cite{lagemann2021deep} and~\cite{Reddy2025}. Because the root-mean-square error weights large residuals more strongly, it exceeds the corresponding averaged endpoint error by a factor that we measure to lie between $1.15$ and $1.4$ for most cases. % >>> CHANGED 2026-10-21: reader half-pixel fix; comparisons re-derived from the rescored rows above.
\rev{Evaluated with the matching statistic on the same test split, $\LIMA{^{3,0}}$ gives lower root-mean-square errors than the published $\WIDIM$, PIV-DCNN, PIV-FlowNetS, PIV-FlowNetS-en, PIV-LiteFlowNet, PIV-LiteFlowNet-en and OF+STE entries on all five cases, and than Horn--Schunck on four of the five, JHTDB-channel ($0.073$ versus $0.069$~px) being the exception.} Our \WIDIM{} re-evaluation lies between $0.55$ and $0.81$ times the published \WIDIM{} values on all five cases, which indicates that the classical baseline tabulated in the literature is not representative of a well-tuned multi-pass correlation analysis.

\section*{\RAFT{} frequency locking and receptive field extent}

\begin{figure}[htbp]
    \centering
    \includegraphics[width=0.6\linewidth, trim=0 0 50 0, clip]{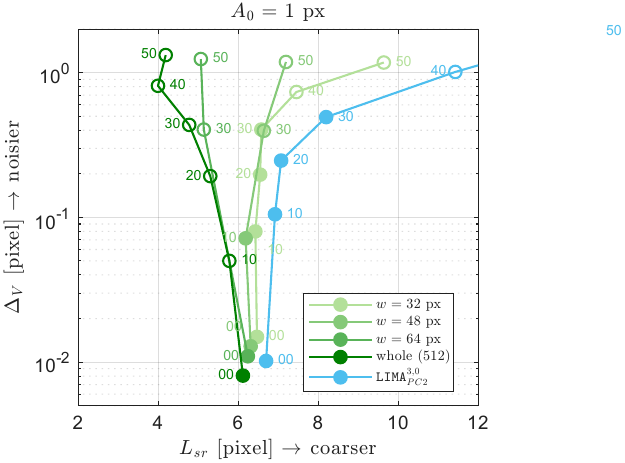}
    \caption{\textbf{Frequency locking in $\RAFT{_{32}}$ is controlled by the
    correlation-window (receptive-field) size.} $\RAFT{_{32}}$ at $A_0=1$~px, its
    all-pairs correlation restricted to non-overlapping windows of side $w=32$, $48$,
    $64$ and $512$~px (whole frame); identical weights, input and output grid
    throughout. Markers give particle-loss levels ($0$, $30$, $50\%$) as
    per-window trajectories in the resolution--noise plane ($L_{sr}$ versus
    $\Delta_V$, log scale); open markers denote a fictitious $L_{sr}$ below the
    window's clean ($0\%$) value. As $w$ grows the trajectories collapse leftward
    to artificially small $L_{sr}$ at nearly unchanged $\Delta_V$, whereas
    $w=32$~px stays faithful.}
    \label{fig:S_windowing}
\end{figure}

In the main text we attribute the frequency-locking artefact of $\RAFT{_{32}}$
to the extent of its effective, end-to-end receptive field rather than to the
all-pairs correlation itself. This distinction is not self-evident: each
correlation lookup in $\RAFT$ is in fact local, sampling a radius of four
displacements at every level of the correlation pyramid and therefore reaching
at most $\pm32$~px for $\RAFT{_{32}}$. The global character of the receptive
field arises only through the recurrent refinement, which couples the per-pixels
estimate across the whole field over its successive updates. To test this
directly, we evaluated $\RAFT{_{32}}$ with its all-pairs correlation restricted
to non-overlapping square windows of side $w = 32, 48$ and $64$~px and,
separately, over the entire $512\times128$~px frame ($w=512$); the response
saturates for larger windows, which coincide with the whole frame. The network
weights, the input image pairs, their normalisation, and the output grid are
identical across all four cases; only the spatial extent over which the
correlation and its recurrent updates operate is varied. Each configuration is
condensed, exactly as in the main text, into a single point in the
resolution--noise plane (equivalent filter length $L_{sr}$ versus random error
$\Delta_V$), and the out-of-plane particle-loss levels are connected into a
per-window trajectory (Supplementary Fig.~\ref{fig:S_windowing}).

At clean conditions the four window sizes coincide, confirming that windowing
does not itself alter the reconstruction when correspondence is intact. Under
particle loss they separate along the resolution axis alone: the random error
$\Delta_V$ is comparable across window sizes, but $L_{sr}$ moves progressively
leftward \rev{towards a spuriously sharper resolution} as the window grows. For the whole frame
$L_{sr}$ falls below its own clean value, the frequency-locking signature (open
symbols), and the deviation deepens with particle loss; the smallest window
($w=32$~px) instead retains a faithful, monotonically coarsening $L_{sr}$, with
$w=64$ interpolating between the two. The leftward
collapse reflects the over-unity amplitude response that the whole-frame
correlation develops at short wavelengths under decorrelation, which the
filter-length transform reads as a fictitiously fine resolution; because
$\Delta_V$ is unchanged, windowing does not make the reconstruction more
accurate, it only removes the spurious sharpening.

Since the network, its weights and the input are held fixed, the only variable
is the extent over which the correlation is coupled. The monotonic dependence of
the artefact on this extent identifies the effective receptive field, not the
per-lookup correlation reach, as its origin, and shows that bounding the
receptive field is sufficient to suppress it. This is the mechanism by which the
coarse-to-fine $\LIMA{^{3}}$ and the finite interrogation window of $\WIDIM$
avoid frequency locking, and it underlies the design principle advanced in the
main text: for dense, full-resolution correlation the effective end-to-end
receptive field must be bounded to the physically relevant displacement range.

\newpage

\section*{Supplementary Tables}

\begin{table}[htbp]
\footnotesize
\centering
\caption{
\rev{Complete $2{\times}2{\times}2{\times}2$ factorial ablation of encoder
depth, decoder receptive field, feature projection, and
correlation-volume normalisation. All sixteen
variants were trained from scratch at full resolution with base learning rate
$2\times10^{-4}$, reduced on plateau until a common floor of $3.1\times10^{-6}$
(151--240~epochs). Six-level variants use a softsign bound on the displacement
update of the decoder, three-level variants a hard bound. Accuracy
was evaluated on the DNS Carlier dataset~\cite{CarlierDNS}.
$\langle\varepsilon\rangle$ and $\sigma$ denote the mean and standard
deviation of the displacement error; $\langle\varepsilon_\zeta\rangle$
and $\sigma_\zeta$ denote the corresponding displacement-curl mean error and standard deviation, in units of $10^{-2}$. Bold values mark the lowest reported point estimate in each column.}
}
\label{tab:ablation_factorial_supp}

{\footnotesize
\begin{tabular}{c c c c c c c c c}
\toprule
Variant &
Encoder &
Decoder RF &
Projector &
Norm. &
$\langle\varepsilon\rangle$ &
$\sigma$ &
$\langle\varepsilon_\zeta\rangle$ &
$\sigma_\zeta$ \\
&
levels &
[px] &
&
&
[px] &
[px] &
[$10^{-2}$] &
[$10^{-2}$] \\
\midrule

% >>> CHANGED 2026-10-29: re-scored against the re-registered 2D-turb ground truth (2026-10-21) with the
% >>>          displacement curl scored against the exact ground-truth curl at the vector positions
% >>>          (score_curl.m); validate/ablation16_v5_L0lr2e4c_turb2d.csv.
v01 & 6 & 15 & no  & no  & \rev{0.1045} & \rev{0.0672} & \rev{3.465} & \rev{2.942} \\
v02 & 6 & 15 & no  & yes & \rev{0.0783} & \rev{0.0495} & \rev{2.860} & \rev{2.462} \\
v03 & 6 & 15 & yes & no  & \rev{0.0870} & \rev{0.0544} & \rev{2.977} & \rev{2.559} \\
v04 & 6 & 15 & yes & yes & \rev{0.0790} & \rev{0.0497} & \rev{2.895} & \rev{2.479} \\
v05 & 6 & 67 & no  & no  & \rev{0.1017} & \rev{0.0634} & \rev{3.439} & \rev{2.973} \\
v06 & 6 & 67 & no  & yes & \rev{0.0814} & \rev{0.0533} & \rev{3.037} & \rev{2.692} \\
v07 & 6 & 67 & yes & no  & \rev{0.0959} & \rev{0.0618} & \rev{3.352} & \rev{2.913} \\
v08 & 6 & 67 & yes & yes & \rev{0.0815} & \rev{0.0523} & \rev{3.079} & \rev{2.681} \\
\midrule
v09 & 3 & 15 & no  & no  & \rev{\textbf{0.0775}} & \rev{0.0484} & \rev{2.853} & \rev{2.462} \\
v10 & 3 & 15 & no  & yes & \rev{0.0782} & \rev{0.0489} & \rev{2.883} & \rev{2.456} \\
v11 & 3 & 15 & yes & no  & \rev{0.0785} & \rev{\textbf{0.0475}} & \rev{\textbf{2.741}} & \rev{\textbf{2.333}} \\
v12 & 3 & 15 & yes & yes & \rev{0.0776} & \rev{0.0482} & \rev{2.865} & \rev{2.441} \\
v13 & 3 & 67 & no  & no  & \rev{0.1005} & \rev{0.0655} & \rev{3.496} & \rev{3.028} \\
v14 & 3 & 67 & no  & yes & \rev{0.0792} & \rev{0.0510} & \rev{2.970} & \rev{2.622} \\
v15 & 3 & 67 & yes & no  & \rev{0.0934} & \rev{0.0574} & \rev{3.192} & \rev{2.769} \\
v16 & 3 & 67 & yes & yes & \rev{0.0819} & \rev{0.0524} & \rev{3.100} & \rev{2.698} \\

\bottomrule
\end{tabular}
}
\end{table}

\section*{Supplementary Figures}

\begin{figure}[htbp!]
    \centering
    \includegraphics[width=0.96\linewidth, trim=0 25 0 0, clip]{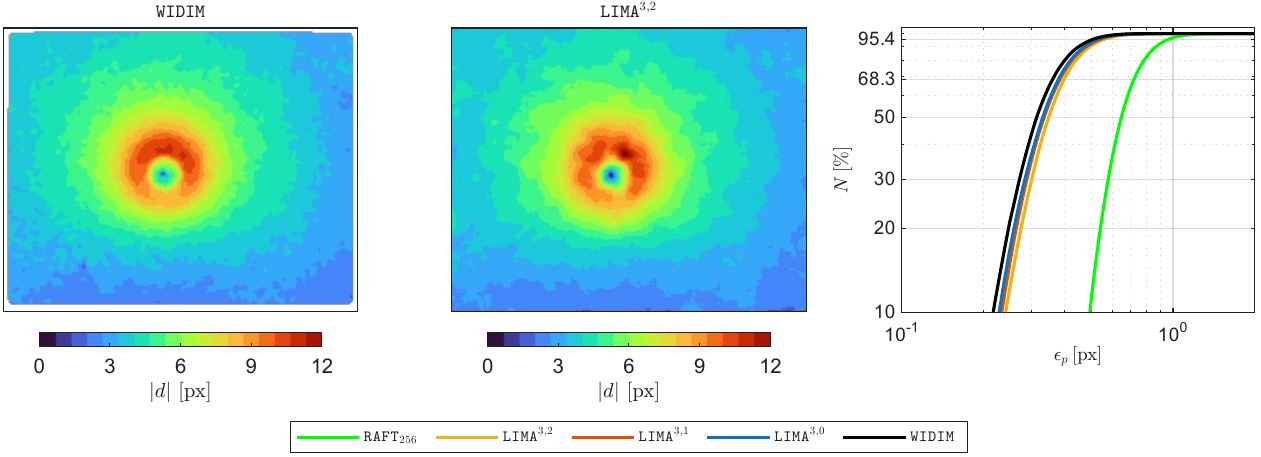}
    \includegraphics[width=0.96\linewidth, trim=0 25 0 10, clip]{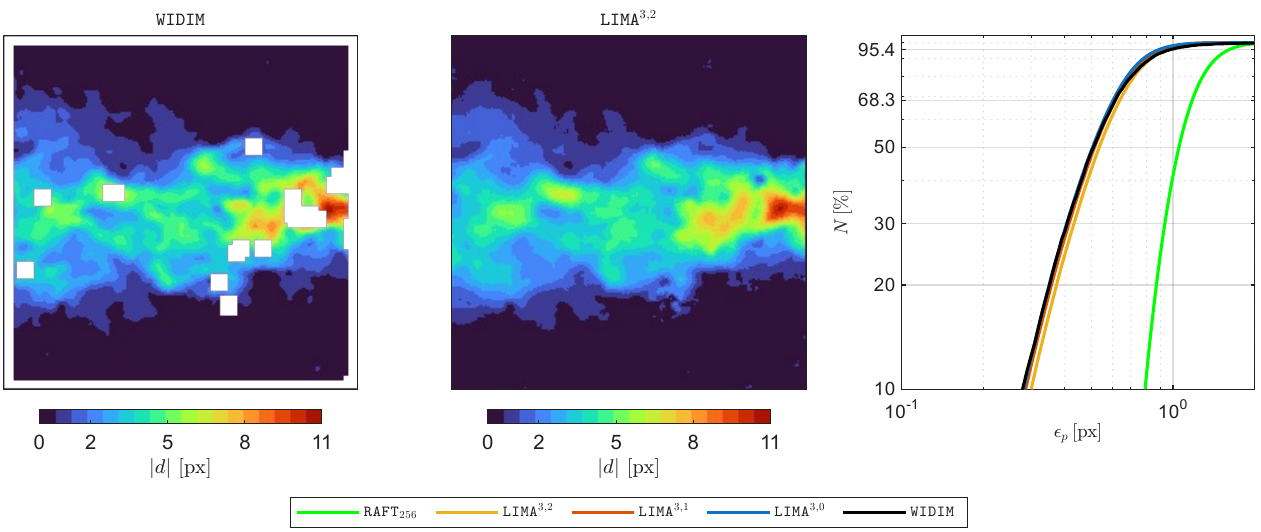}
    \includegraphics[width=0.96\linewidth, trim=0 25 0 10, clip]{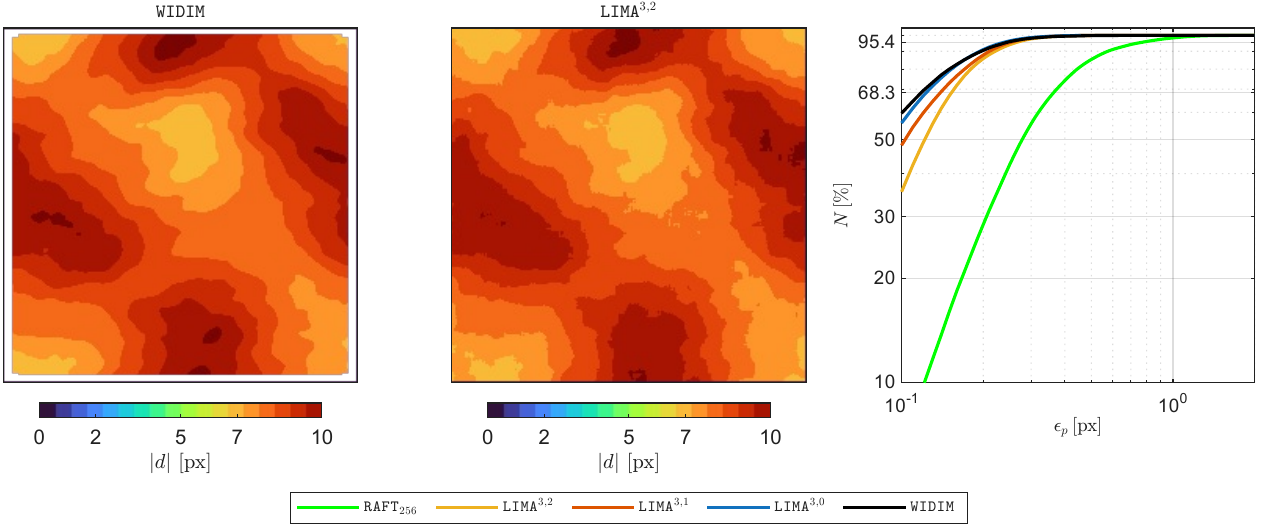}
    \includegraphics[width=0.96\linewidth, trim=0 0 0 10, clip]{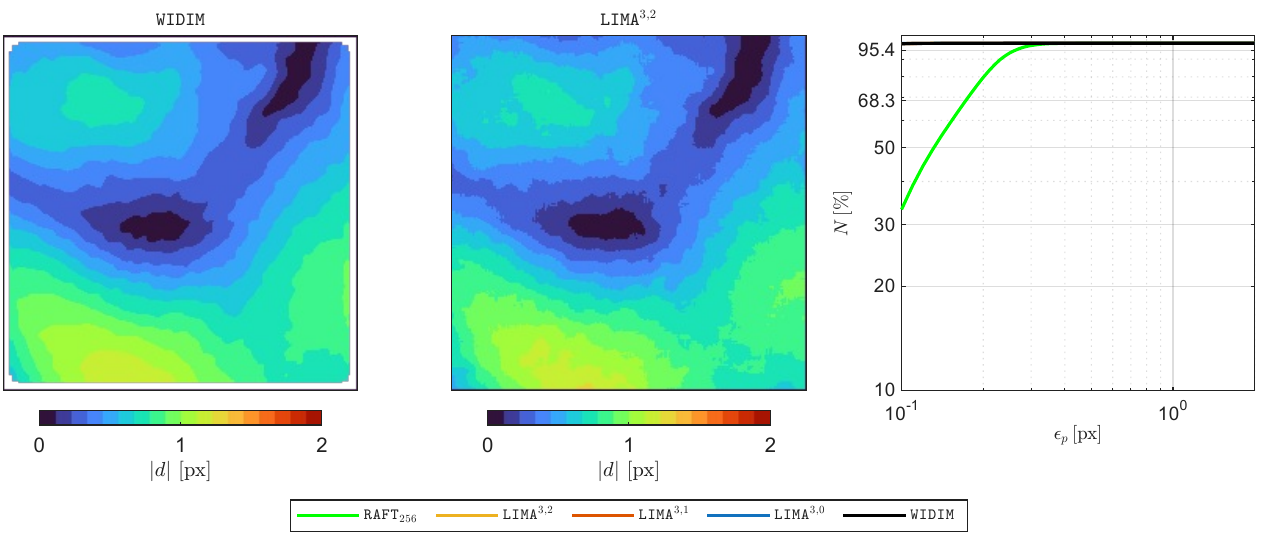}
    \caption{PIV Challenge benchmark, group 1.
    Left: displacement magnitude from $\WIDIM$. Middle: displacement magnitude
    from $\LIMA{^{3,2}}$. Right: histograms of disparity error $\epsilon_p$
    for $\WIDIM$,  the three $\LIMA$ variants, and \RAFT{$_{256}$}.
    Rows: PIV Challenge 1 case A; Challenge 2 case A; Challenge 3 case A2;
    Challenge 3 case A3.}
    \label{fig:Exp2_supp}
\end{figure}

\begin{figure}[htbp]
    \centering
    \includegraphics[width=0.96\linewidth, trim=0 25 0 0, clip]{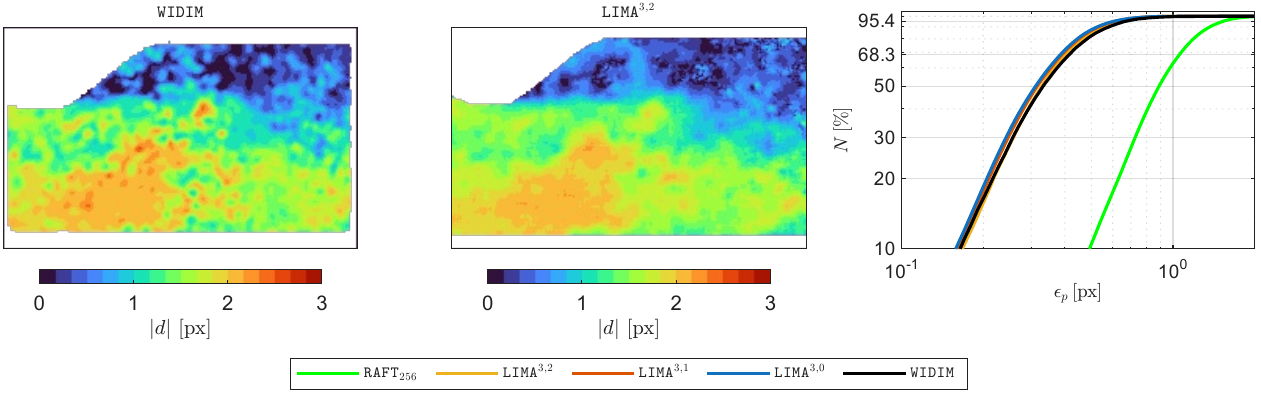}
    \includegraphics[width=0.96\linewidth, trim=0 25 0 10, clip]{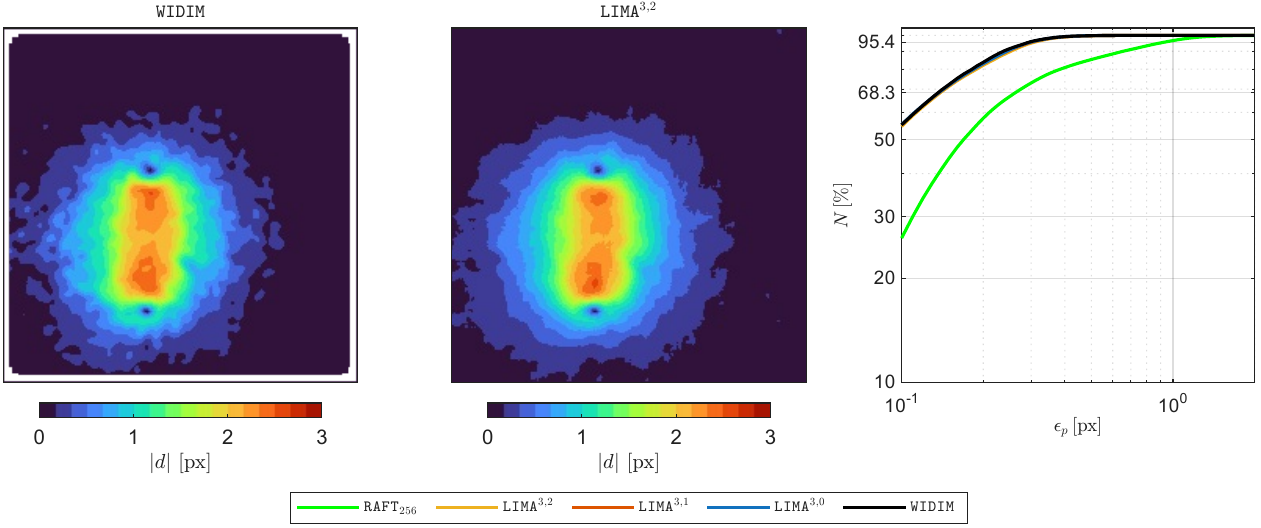}
    \includegraphics[width=0.96\linewidth, trim=0 25 0 10, clip]{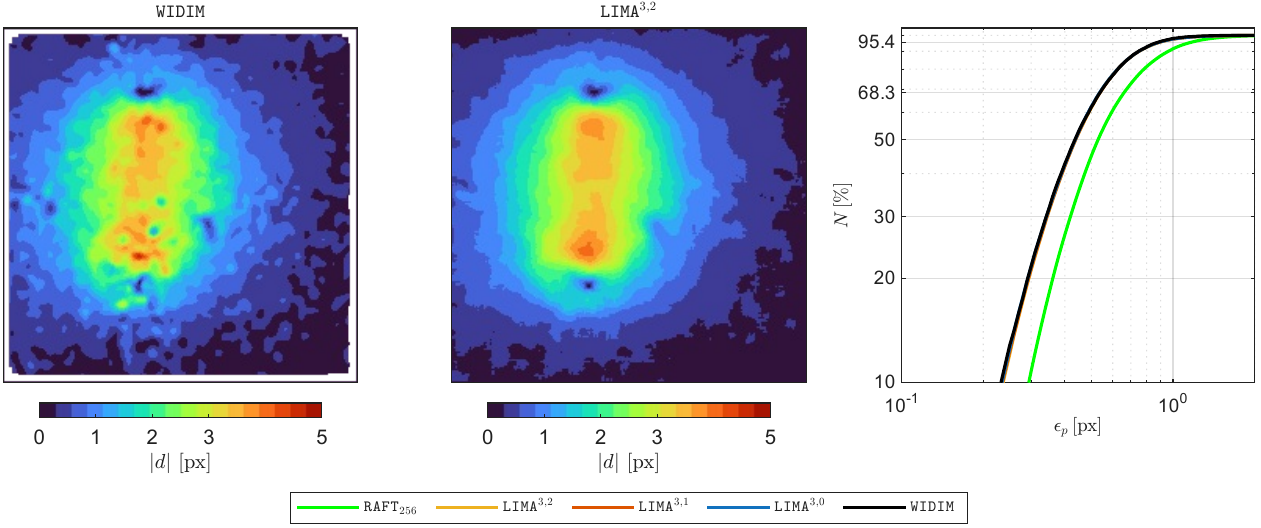}
    \includegraphics[width=0.96\linewidth, trim=0 0 0 10, clip]{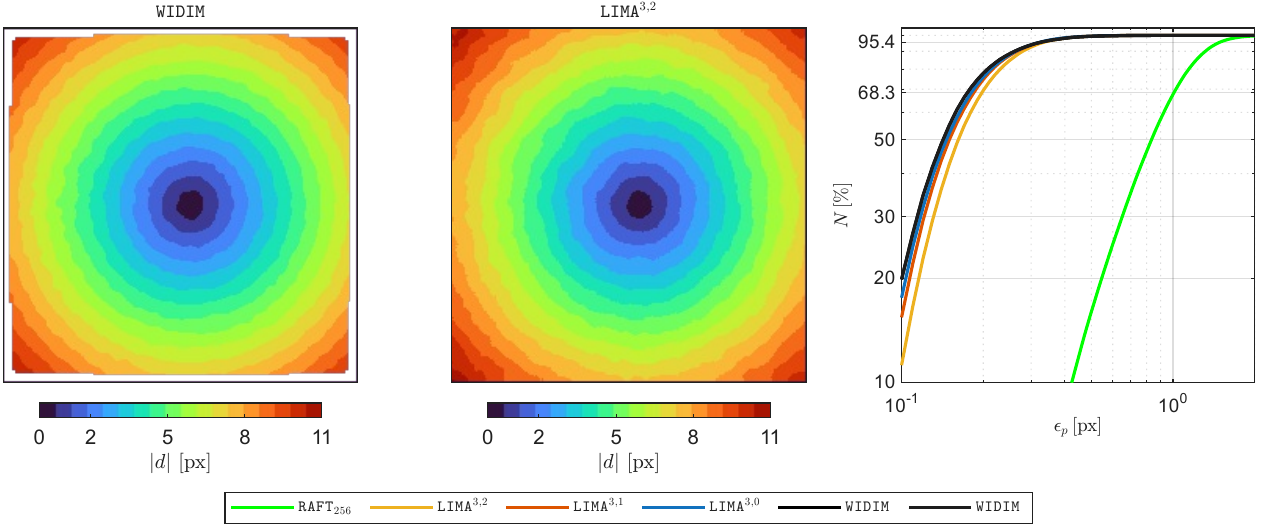}
    \caption{PIV Challenge benchmark, group 2.
    Panels and methods as in Supplementary Figure~\ref{fig:Exp2_supp}.
    Rows: PIV Challenge 4 case B; Challenge 4 case E camera 1;
    Challenge 4 case E \rev{camera 3}; Challenge 4 case F.}
    \label{fig:Exp3_supp}
\end{figure}

\begin{figure}[htbp]
    \centering
    \includegraphics[width=0.96\linewidth, trim=0 25 0 0, clip]{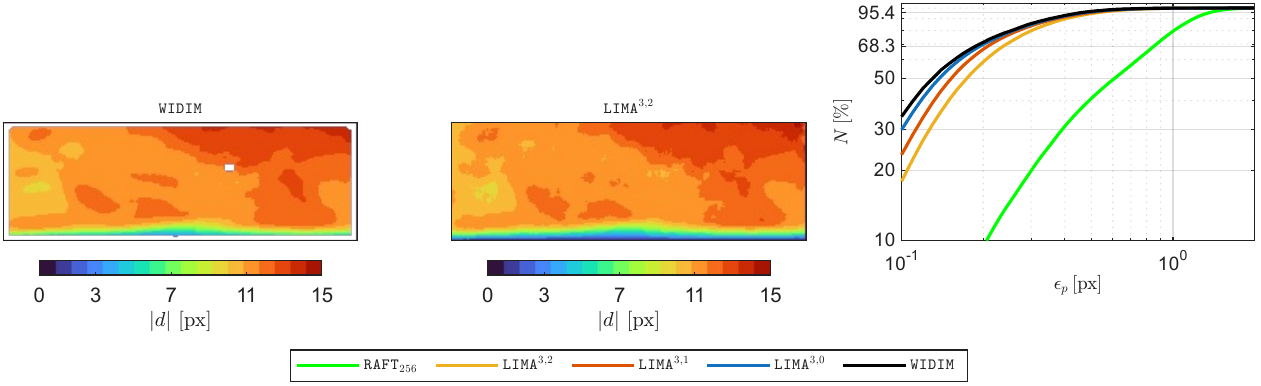}
        \includegraphics[width=0.96\linewidth, trim=0 25 0 0, clip]{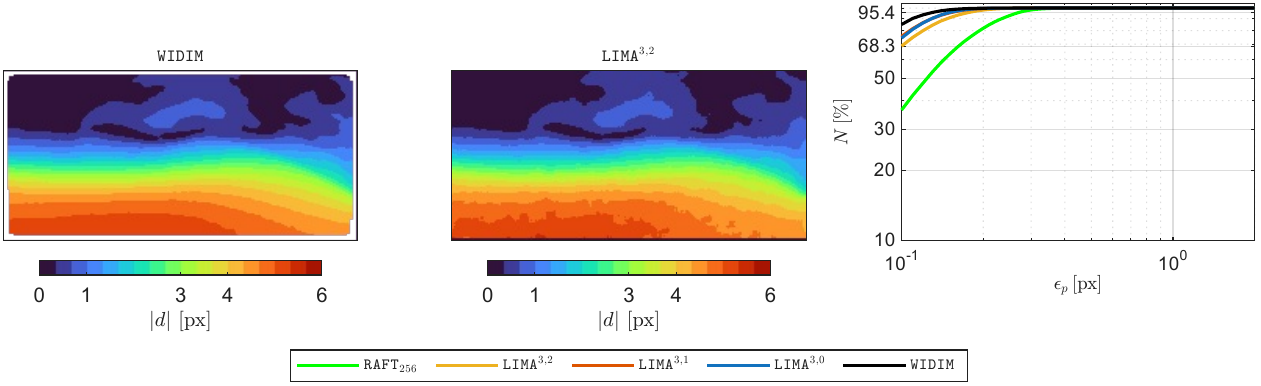}
    \includegraphics[width=0.96\linewidth, trim=0 25 0 0, clip]{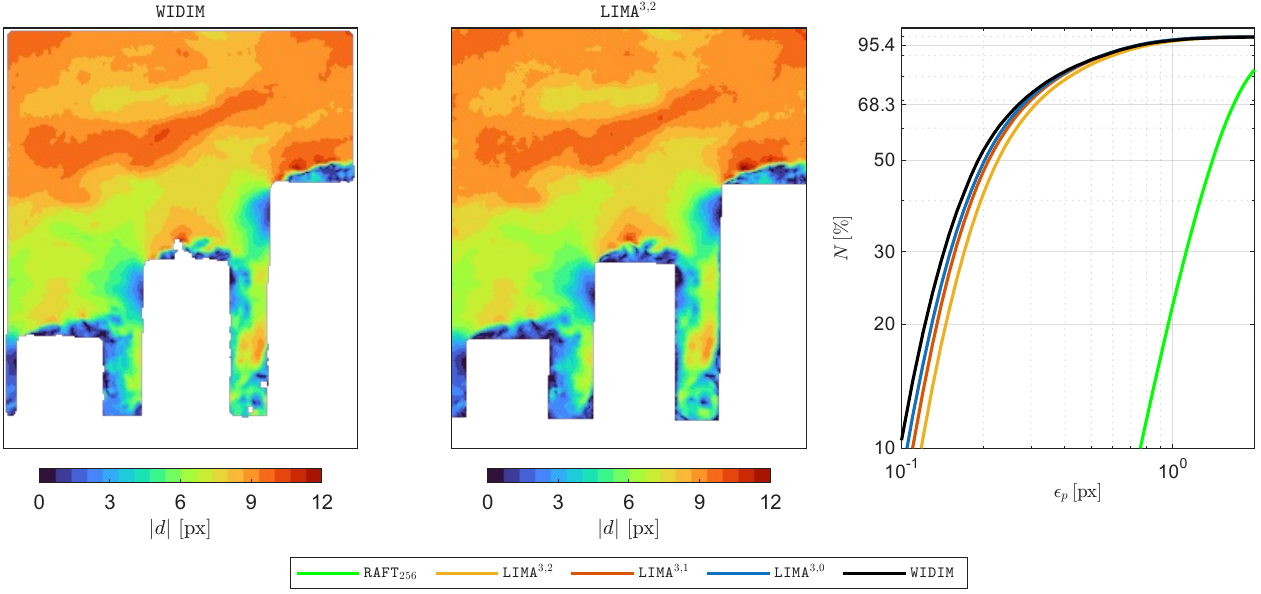}
\includegraphics[width=0.96\linewidth, trim=0 0 0 10, clip]{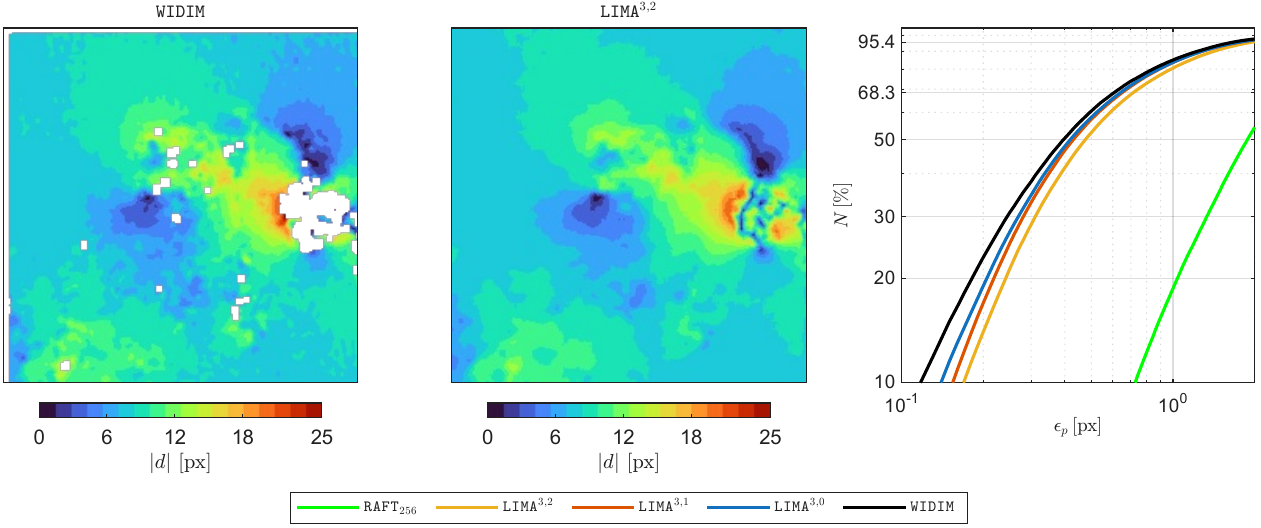}
    \caption{Additional experimental flows. Panels and methods as in
    Supplementary Figure~\ref{fig:Exp2_supp}. Rows: \rev{PIV Challenge 2 case B; PIV Challenge 3 case B; urban canopy (buildings); flow past a robotic fish.}}
    \label{fig:Exp4_supp}
\end{figure}

\begin{figure}[htbp]
    \centering
    \includegraphics[width=0.92\linewidth]{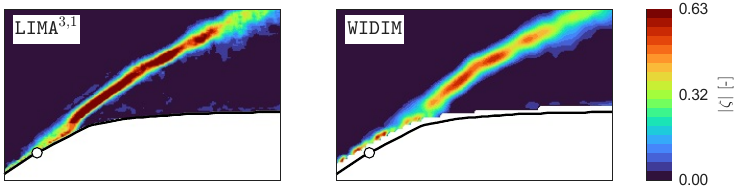}
% >>> CHANGED 2026-10-29: render-1.0 LIMA^{3,1}, ten pairs; roof sentence removed (user: cable car only).
    \caption{\rev{\textbf{Separated shear layer at the cable-car nose, close to the wall.}
    Magnitude of the curl of the displacement field averaged over the ten image pairs, for $\LIMA{^{3,1}}$
    (left) and $\WIDIM{}$ (right). The $\LIMA{^{3,1}}$ high curl is a narrow line that reaches the surface, whereas the $\WIDIM{}$ ridge is broad and diffuse.}}
    \label{fig:separation_supp}
\end{figure}

\begin{figure}[htbp]
    \centering
    \includegraphics[width=1\linewidth]{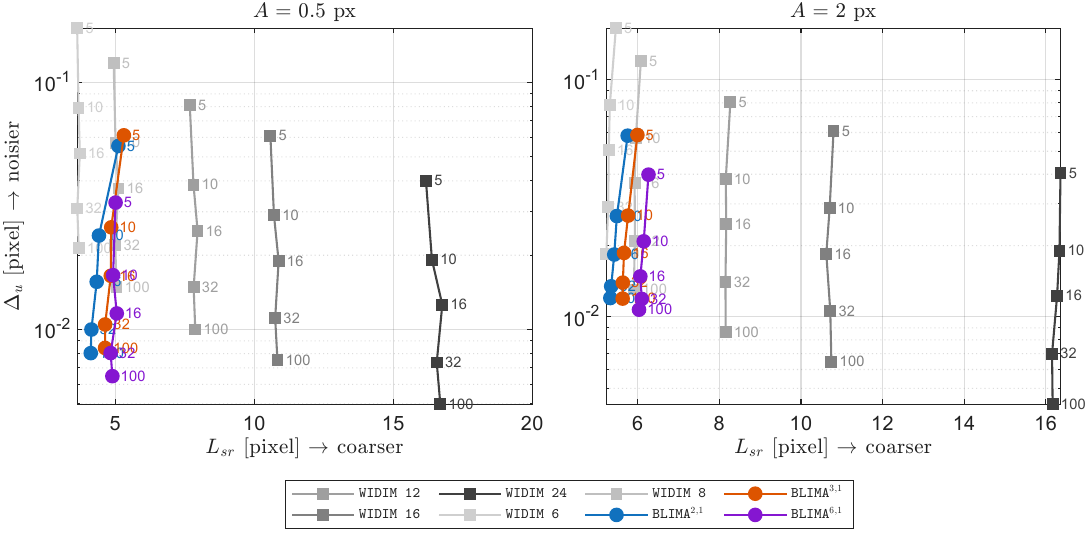}
    \caption{Supplementary  network  depth sweep under image-SNR degradation for \BLIMA{}, a variant of \LIMA{} trained on BOS data as done in \cite{mucignat2023lightweight}. Resolution-versus-noise trade-off for the sinusoidal BOS benchmark at
$A=0.5$~px (left) and $A=2$~px (right). The horizontal axis reports the
equivalent filter length $L_{sr}$, with larger values corresponding to
coarser spatial resolution; the vertical axis reports the random
displacement error. Marker labels indicate image SNR levels. Unlike the PIV
particle-loss sweep, BOS degradation is imposed through image SNR. The \BLIMA{} variants provide a consistency
check for the receptive-field budget: shallower encoders are favoured when the
physical displacement scale is smaller.}
\label{fig:bos_pareto_supp}
\end{figure}

%%%%%%%%%%%%%%%%%%%%%%%%%%%%%%%%%%%%%%%%%%%%%%%%%%%%%%%%%%%%%%%%%%%%%
%% END OF SUPPLEMENTARY METHODS BLOCK                                %
%%%%%%%%%%%%%%%%%%%%%%%%%%%%%%%%%%%%%%%%%%%%%%%%%%%%%%%%%%%%%%%%%%%%%

\end{document}